\documentclass[journal]{IEEEtran}

\usepackage{amssymb}
\usepackage{color}
\usepackage[fleqn]{amsmath}
\usepackage{graphicx}
\usepackage{float}
\usepackage{multirow}
\usepackage{caption}
\usepackage{array}
\usepackage{rotating}
\usepackage{arydshln}
\usepackage{makeidx}
\usepackage{subfigure}

\usepackage{multicol}

\usepackage{stfloats}

\usepackage{subeqnarray}
\usepackage{cases}

\usepackage[colorlinks,
            linkcolor=red,
            anchorcolor=blue,
            citecolor=green
            ]{hyperref}

\usepackage{threeparttable}

\makeindex

\newcommand{\MYnextline}[2]{\begin{tabular}{@{}#1@{}}#2\end{tabular}}

\newcommand {\Scale}[0]{0.90}

\makeatletter

\ifCLASSINFOpdf
\else
\fi
\begin{document}
%
\title{A Unified Gender-Aware Age Estimation}
%
%
%

\author{Qing~Tian, Songcan~Chen$^{*}$, and Xiaoyang Tan
\thanks{Qing~Tian, Songcan~Chen and Xiaoyang Tan are with the College of Computer Science and Technology, Nanjing University of Aeronautics and Astronautics, Nanjing 210016, China
(e-mail: tianqing@nuaa.edu.cn (Q. Tian); s.chen@nuaa.edu.cn (S. Chen, Corresponding Author); x.tan@nuaa.edu.cn (X. Tan)).}
\thanks{Manuscript received on Mar 11, 2016.}}

%
%

\markboth{IEEE Transactions on Cybernetics}%
{Shell \MakeLowercase{\textit{et al.}}: Bare Demo of IEEEtran.cls for Journals}
%




\maketitle

\begin{abstract}
Human age estimation has attracted increasing researches due to its wide applicability in such as security monitoring and advertisement recommendation. Although a variety of methods have been proposed, most of them focus only on the age-specific facial appearance. However, biological researches have shown that not only gender but also the aging difference between the male and the female inevitably affect the age estimation. To our knowledge, so far there have been two methods that have concerned the gender factor. The first is a sequential method which first classifies the gender and then performs age estimation respectively for classified male and female. Although it promotes age estimation performance because of its consideration on the gender semantic difference, an accumulation risk of estimation errors is unavoidable. To overcome drawbacks of the sequential strategy, the second is to regress the age appended with the gender by concatenating their labels as two dimensional output using Partial Least Squares (PLS). Although leading to promotion of age estimation performance, such a concatenation not only likely confuses the semantics between the gender and age, but also ignores the aging discrepancy between the male and the female. In order to overcome their shortcomings, in this paper we propose a unified framework to perform gender-aware age estimation. The proposed method considers and utilizes not only the semantic relationship between the gender and the age, but also the aging discrepancy between the male and the female. Finally, experimental results demonstrate not only the superiority of our method in performance, but also its good interpretability in revealing the aging discrepancy.
\end{abstract}

\begin{IEEEkeywords}
Age estimation; Gender-aware age estimation; Threshold-based ordinal regression; Support vector ordinal regression
\end{IEEEkeywords}

%
\IEEEpeerreviewmaketitle

\section{Introduction} \label{sec:introduction}
\IEEEPARstart{H}{uman} face conveys various information, such as the gender, age, race and expression, etc., in which human age estimation has attracted increasing attention in recent years due to its wide applications in recommendation systems \cite{linoff2011data}, \cite{raab2012customer}, security access control \cite{LARR_guo2008image}, \cite{wu2013attribute}, biometrics \cite{van2004dependency}, \cite{patterson2007aspects} and entertainment \cite{chen2013travel}, \cite{dibeklioglu2015recognition}, etc.

To perform age estimation according to human facial appearance, a variety of methods have been developed. These methods can be roughly grouped into three categories: classification-based (e.g., \cite{lanitis2004comparing}, \cite{geng2013facial}, \cite{ueki2006subspace}, \cite{alnajar2012learning}, \cite{sai2015facial}, \cite{alnajar2014expression}, \cite{dibeklioglu2015combining}), regression-based (e.g., \cite{lanitis2002toward}, \cite{fu2007estimating}, \cite{luu2009age}, \cite{yan2007ranking}, \cite{yan2007learning}, \cite{geng2007automatic}, \cite{fu2008human}, \cite{chang2011ordinal}, \cite{ni2011web}, \cite{li2012learning_ICPRmetric}, \cite{li2012learning_learnforfeature}, \cite{chao2013facial}), and their hybrid methods (e.g., \cite{LARR_guo2008image}, \cite{kohli2013hierarchical}). When treating each age as an ordinary class, we can perform age estimation using the multi-class classification framework. Along this line, artificial neural networks (ANN) \cite{lanitis2004comparing}, conditional probability neural networks (CPNN) \cite{geng2013facial}, Gaussian mixture models \cite{ueki2006subspace}, and extreme learning machines (ELM) \cite{sai2015facial} have been successively employed for age estimation. More recently, \cite{alnajar2014expression} proposed an expression-insensitive age estimation method. \cite{dibeklioglu2015combining} performed age estimation by incorporating facial dynamics together with the appearance information. Actually, age estimation is more of a regression problem rather than classification due to its characteristics of continuity and monotonicity. Along this line, quadratic function \cite{lanitis2002toward}, \cite{fu2008human}, multiple linear regression \cite{fu2007estimating}, $\xi$-SVR \cite{luu2009age}, SDP regressors \cite{yan2007ranking, yan2007learning}, aging pattern subspace (AGES) \cite{geng2007automatic}, multi-instance regressor \cite{ni2011web}, and KNN-SVR \cite{chao2013facial} have been successively proposed to perform human age regression. Besides the above age estimations based on classification or regression, a hybrid strategy has also been adopted. For example, \cite{LARR_guo2008image} established a so-called locally adjusted robust regression (LARR) to predict human age by combining a series of classifiers and regressors.

Although the aforementioned methods are applicable for estimating human age, most of them focus only on age-specific facial appearance. Actually, biological researches have shown that the male and the female are aging differently \cite{holland1975adaptation}. That is, not only the gender but also the aging difference between the male and the female inevitably affect the age estimation. To perform age estimation with concerning the gender factor, \cite{guo2010human} and \cite{lakshmiprabha2011age} proposed a sequential method which first classifies the gender and then performs age estimation respectively for classified male and female. Although such a sequential method promotes age estimation performance because of its consideration on the gender semantic difference, an accumulation risk of estimation errors is unavoidable. Another work is \cite{guo2014framework}, which adopted the Partial Least Squares (PLS) to regress human age together with gender by concatenating their labels as a two-dimensional regression output, and achieved state-of-the-art accuracies on several benchmark aging datasets. Although leading to promotion of age estimation performance, such a concatenation not only likely confuses the semantics between the gender and age, but also ignores the aging discrepancy between the male and the female.

To overcome the above drawbacks of existing methods, in this paper we propose a unified framework to perform gender-aware age estimation by dividing the entire gender space into male and female subspaces while making gender-aware age estimation in respective male and female subspaces. Finally, through experiments, we demonstrate the superiority of the proposed method in age estimation accuracy, and also explore the aging discrepancy between the male and the female.

The rest of this paper is organized as follows. In Section \ref{sec:related work}, we review related works. In Sections \ref{sec:proposed-methodology} and \ref{sec:experiment}, we present and experimentally evaluate the proposed method, respectively. Finally, Section \ref{sec:conclusions} concludes this paper.

\section{Related Work} \label{sec:related work}
\begin{figure}
  \centering
  \subfigure[]{
    \label{hierarchical-1-flow} 
    \includegraphics[width=0.80\linewidth, height=1.3in]{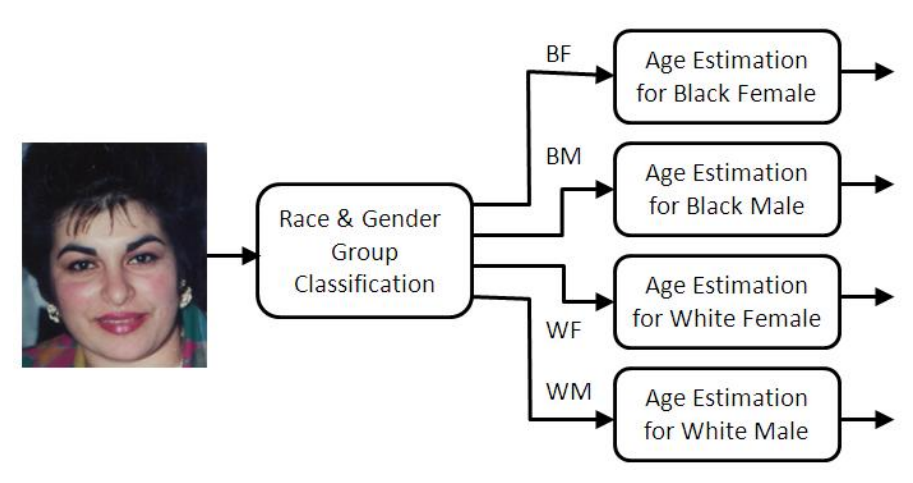}}
  \subfigure[]{
    \label{hierarchical-1-flow} 
    \includegraphics[width=0.80\linewidth, height=1.4in]{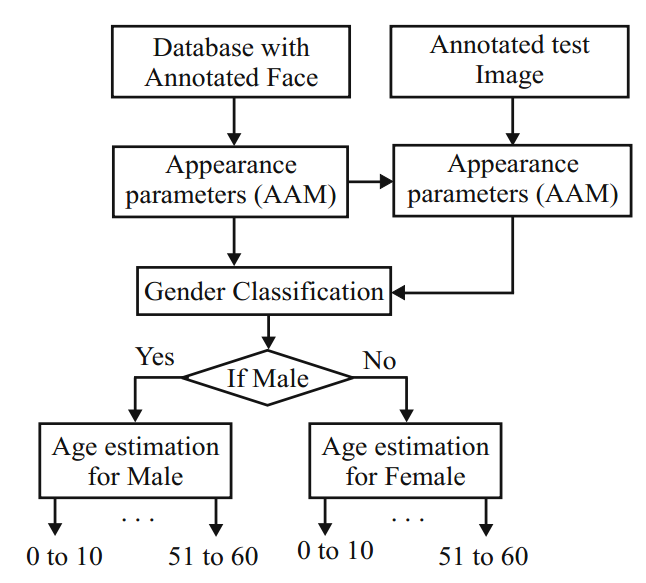}}
  \caption{The flowchart of \cite{guo2010human} (a) and \cite{lakshmiprabha2011age} (b).}
  \label{fig:hierarchical-flow} 
\end{figure}

To our knowledge, so far there have been two methods that have concerned the gender factor. The first is a sequential method which first classifies the gender and then performs age estimation respectively for classified male and female \cite{guo2010human}, \cite{lakshmiprabha2011age}. Concretely, to factor out the gender variation in age estimation, \cite{guo2010human} and \cite{lakshmiprabha2011age} took a two-step strategy, i.e., perform gender identification and then male- or female-specific age estimation. Their flowcharts are shown in Figure \ref{fig:hierarchical-flow}.
The flowcharts show that such a two-step strategy requires to make gender identification before age estimation, incurring the latter's performance extremely suffers from the errors made in gender estimation. To overcome drawbacks of the aforementioned sequential method, the second method simultaneously regresses the age appended with gender by concatenating their labels as a two-dimensional regression output \cite{guo2014framework} using Partial Least Squares (PLS) \cite{geladi1986partial}. For PLS, it typically aims to maximize the correlation between input, $X$, and output, $Y$, in the projected spaces characterized by $w$ and $c$, through optimizing the below objective function:
\begin{equation}\label{eq:pls}
    \scalebox{\Scale}
    {$
\begin{split}
& max_{|w|=|c|=1} \; cov(w^TX, c^TY).
\end{split}
    $}
\end{equation}
It can be seen from Eq. \eqref{eq:pls} that although the gender and age output labels are concatenated in $Y$ as a whole regression target and learned in a joint way in \cite{guo2014framework}, not only the semantics between the gender and age is confused, but also the aging discrepancy between the male and the female is ignored, due to lack of explicit constraint or regularization regarding such semantic relations in its objective function. To overcome the drawbacks of aforementioned methods, in the next section we propose a unified framework to perform gender-aware age estimation by explicitly preserving and utilizing not only the semantic relationship between the gender and the age, but also the aging discrepancy between the male and the female.

\section{Proposed Methodology} \label{sec:proposed-methodology}
\subsection{A Unified Framework for Gender-Aware Age Estimation (GenAge)} \label{sec:GenAge}
\begin{figure}[htdp!]
  \centering
  \includegraphics[width=0.95\linewidth]{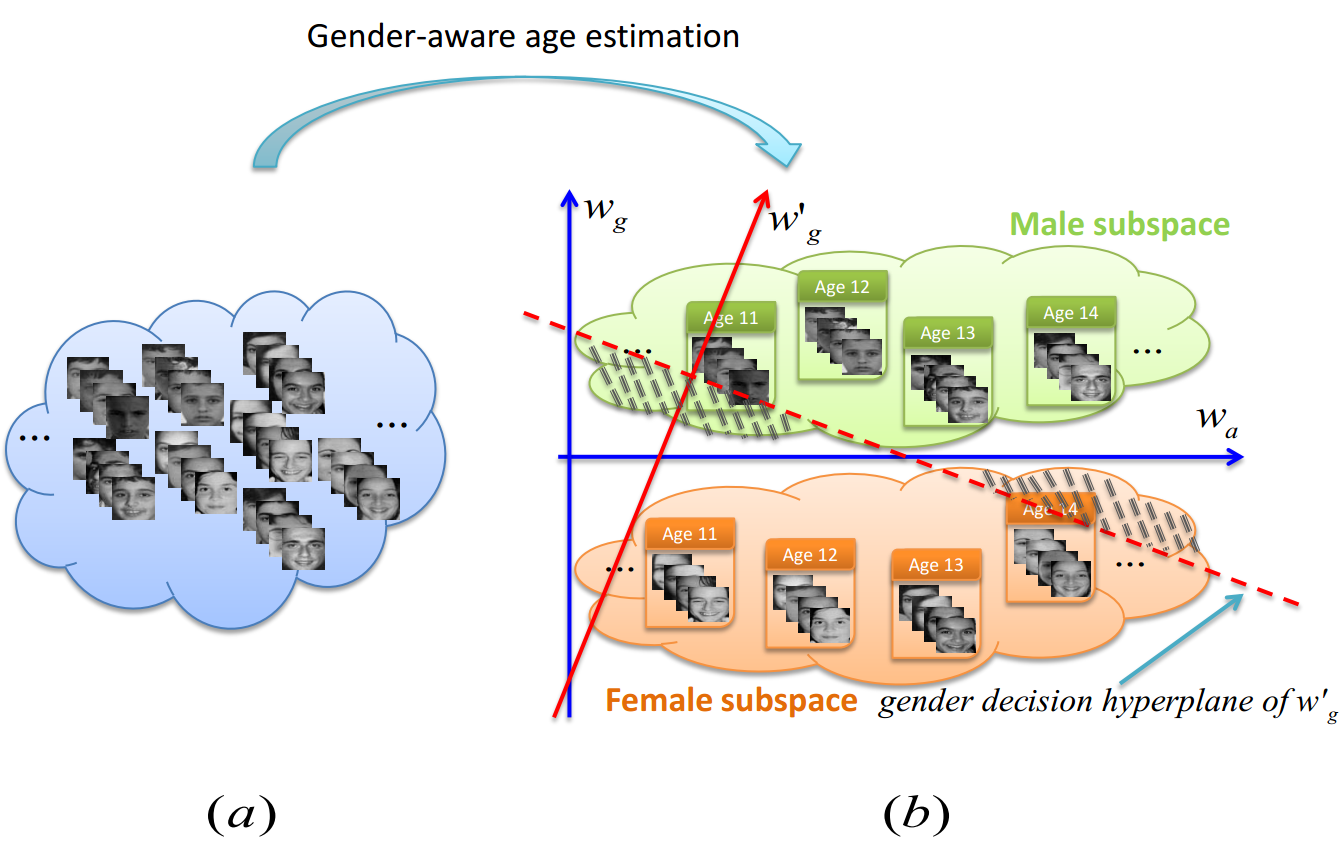}\\
  \caption{Illustration of the proposed gender-aware age estimation, from the original sample space (a) into gender-aware age estimation space (b). $w_g$ ($w_g^{'}$) and $w_a$ denote the gender discriminant direction and human aging direction, respectively.}\label{fig:flowchart-of-GenAge}
\end{figure}
As shown in Figure \ref{fig:flowchart-of-GenAge}, let $w_a$ denote the aging direction, and $w_g$ or $w_g^{'}$ indicates the gender discriminant direction. The samples distribute monotonously and orderly in terms of their age labels along $w_a$ \cite{geng2007automatic}, \cite{tian2015ca}. When $w_g^{'}$ is chosen as the gender discriminant direction, the male and female gender subspaces will cross the gender decision hyperplane of $w_g^{'}$ (see the area shaded by oblique lines in Figure \ref{fig:flowchart-of-GenAge}(b)), implying that a severe gender misclassification. By contrast, if we choose $w_g$, which is nearly orthogonal to $w_a$, as the gender discriminant direction, the male and female gender subspaces will more clearly distribute on two sides of its gender decision hyperplane. For the sake of formulation, we advocate to use a squared inner-product term
\begin{equation}\label{eq:orthogonalTerm}
    \scalebox{\Scale}
    {$
\begin{split}
& \mathcal{R}(w_g, w_a):= (w_g^Tw_a)^2
\end{split}
    $}
\end{equation}
between $w_g$ and $w_a$ to depict their space relation. And according to the analysis above, term \eqref{eq:orthogonalTerm} should be minimized to enforce the near-orthogonality between $w_g$ and $w_a$. With Eq. \eqref{eq:orthogonalTerm} as an explicit regularization term, we come to construct a unified \emph{gender-aware age estimation} (GenAge) framework as follows:
\begin{equation}\label{eq:GenAge-framework}
    \scalebox{\Scale}
    {$
\begin{split}
& min_{\{w_{g}, w_{a}\}} \; \mathcal{L}_{g}(w_g;X) + \mathcal{L}_{a}^{male}(w_a;X) + \mathcal{L}_{a}^{female}(w_a;X) \\
& \quad \qquad \qquad + \lambda \cdot\mathcal{R}(w_g, w_a),
\end{split}
    $}
\end{equation}
where $X$ represents the training data, $\mathcal{L}_{g}(w_g;X)$ stands for loss function for gender classification, $\mathcal{L}_{a}^{male}(w_a;X)$ and $\mathcal{L}_{a}^{female}(w_a;X)$ are corresponding loss functions of age estimation regarding the male and the female, respectively, and $\lambda$ is a nonnegative trade-off parameter. By the above modelling manner, the semantic relationship between the gender and the age, as well as the aging discrepancy between the male and the female are successfully incorporated in Eq. \eqref{eq:GenAge-framework}.

\subsection{Exemplification of the GenAge Framework} \label{sec:exemplified-GenAge}
To practically evaluate GenAge framework in Eq. \eqref{eq:GenAge-framework}, in this section we exemplify it. Concretely, assume that there are a total of $N$ training samples, from $K$ ages. For gender classification, in view of its binariness, without loss of generality we take the widely-used SVM as classifier and consequently substitute $\mathcal{L}_{g}(w_g;X)$ with hinge loss. For age estimation, considering the competitive performance of support vector ordinal regression (SVOR) \cite{chu2005new} in ordinal regression\footnote{Note that besides the SVM and SVOR, we can also take other binary classification methods (e.g., logistic regression \cite{anderson1982logistic}) for gender classification and threshold-based ordinal regression methods (e.g., KDLOR \cite{sun2010kernel}) for age estimation, respectively. However, in this work our focus is on exploring the influence of human gender to age estimation, so we do not give other more exemplifications.}, we substitute it for $\mathcal{L}_{a}^{male}(w_a;X)$ and $\mathcal{L}_{a}^{female}(w_a;X)$, respectively. After these substitutions, Eq. \eqref{eq:GenAge-framework} can be rewritten as follows:
\begin{equation}\label{eq:SVM-SVOR-GenAge}
    \scalebox{\Scale}
    {$
\begin{split}
& min_{\{w_g, b_g, w_a, b^m:=\{b^m_k\}_{k=1}^{K}, b^f:=\{b^f_k\}_{k=1}^{K}\}} \\
& \qquad \frac{1}{2}\|w_g\|^2 + \lambda_{1}\sum_{i=1}^{N}max\{0, 1-y^g_i(w^T_gx^g_i+b_g)\} \\
& \qquad \frac{1}{2}\|w_a\|^2 + \lambda_{2}\sum_{k=1}^{K}(\sum_{i=1}^{N^{m}_{k}}(\xi^k_{mi}+\xi^{k*}_{mi})+\sum_{j=1}^{N^{f}_{k}}(\xi^k_{fj}+\xi^{k*}_{fj})) \\
& \qquad + \lambda_{3}(w^T_gw_a)^2\\
& s.t.\\
& \qquad w^T_ax^{k}_{mi} - b^m_k \leq -1 + \xi^k_{mi}, \; \xi^k_{mi} \geq 0, \\
& \qquad w^T_ax^{k*}_{mi} - b^m_{k-1} \leq -1 - \xi^{k*}_{mi}, \; \xi^{k*}_{mi} \geq 0, \\
& \qquad b^m_{k-1} \leq b^m_k, \\
& \qquad w^T_ax^{k}_{fj} - b^f_k \leq -1 + \xi^k_{fj}, \; \xi^k_{fj} \geq 0, \\
& \qquad w^T_ax^{k*}_{fj} - b^f_{k-1} \leq -1 - \xi^{k*}_{fj}, \; \xi^{k*}_{fj} \geq 0, \\
& \qquad b^f_{k-1} \leq b^f_k,
\end{split}
    $}
\end{equation}
where $w_g$ and $b_g$ denote projection vector and bias of SVM, $w_a$ represents weight vector of SVOR, $b^m$ and $b^f$ are age regression thresholds in terms of the male and the female, respectively. $\xi$'s denote slack variables, $x$'s stand for the training samples, and $\lambda_1$, $\lambda_2$ and $\lambda_3$ are nonnegative trade-off parameters.

Eq. \eqref{eq:SVM-SVOR-GenAge} is a bi-convex problem with respect to $\{w_{g}, b_{g}\}$ and $\{w_{a}, b^{m}, b^{f}\}$. Thus, we can take an alternating strategy to optimize it. More precisely, for fixed $\{w_{a}, b^{m}, b^{f}\}$, Eq. \eqref{eq:SVM-SVOR-GenAge} becomes
\begin{equation}\label{eq:SVM&SVM-SVOR-GenAge}
    \scalebox{\Scale}
    {$
\begin{split}
& min_{\{w_{g}, b_{g}\}} \\
& \qquad \frac{1}{2}w_{g}^{T}(\textbf{I}+2\lambda_{3}w_{a}w_{a}^{T})w_{g} + \lambda_{1}\sum_{i=1}^{N}max\{0, 1-y^{g}_{i}(w_{g}^{T}x^{g}_{i}+b_{g})\},
\end{split}
    $}
\end{equation}
where $\textbf{I}$ represents an identity matrix of proper size. The sub-problem \eqref{eq:SVM&SVM-SVOR-GenAge} is convex with respect to $\{w_{g}, b_{g}\}$ and can be similarly solved by the same way as SVM \cite{steinwart2008support}.

In turn, when $\{w_{g}, b_{g}\}$ are obtained, we can fix them and equivalently write Eq. \eqref{eq:SVM-SVOR-GenAge} as
\begin{equation}\label{eq:SVOR&SVM-SVOR-GenAge}
    \scalebox{\Scale}
    {$
\begin{split}
& min_{\{w_{a}, b^{m}, b^{f}\}} \\
& \qquad \frac{1}{2}w_{a}^{T}(\textbf{I} + 2\lambda_{3}w_{g}w_{g}^{T})w_{a} + \lambda_{2}\sum_{k=1}^{K}(\sum_{i=1}^{N^{m}_{k}}(\xi^k_{mi}+\xi^{k*}_{mi})+\sum_{j=1}^{N^{f}_{k}}(\xi^k_{fj}+\xi^{k*}_{fj}))\\
& s.t. \\
& \qquad w^T_ax^{k}_{mi} - b^m_k \leq -1 + \xi^k_{mi}, \; \xi^k_{mi} \geq 0, \\
& \qquad w^T_ax^{k*}_{mi} - b^m_{k-1} \leq -1 - \xi^{k*}_{mi}, \; \xi^{k*}_{mi} \geq 0, \\
& \qquad b^m_{k-1} \leq b^m_k, \\
& \qquad w^T_ax^{k}_{fj} - b^f_k \leq -1 + \xi^k_{fj}, \; \xi^k_{fj} \geq 0, \\
& \qquad w^T_ax^{k*}_{fj} - b^f_{k-1} \leq -1 - \xi^{k*}_{fj}, \; \xi^{k*}_{fj} \geq 0, \\
& \qquad b^f_{k-1} \leq b^f_k,
\end{split}
    $}
\end{equation}
which is similar to the problem (5) in \cite{chu2005new} and can be solved similarly.

To obtain desirable $\{w_{g}, b_{g}, w_{a}, b_{a}, \xi^{(*)}\}$, we repeat the alternating optimization process between Eqs. \eqref{eq:SVM&SVM-SVOR-GenAge} and \eqref{eq:SVOR&SVM-SVOR-GenAge} until it converges\footnote{Note that since both Eqs. \eqref{eq:SVM&SVM-SVOR-GenAge} and \eqref{eq:SVOR&SVM-SVOR-GenAge} are convex, their objective function values decrease with increasing iterations and finally converge to their optimal values, so the objective function value of Eq. \eqref{eq:SVM-SVOR-GenAge} converges as well.}. We summarize the complete optimization algorithm in Table \ref{tab: algorithm SVM-SVOR-GenAge}.
\begin{table}[!ht]
\centering
\caption{An alternating algorithm for Eq. \eqref{eq:SVM-SVOR-GenAge}.}\label{tab: algorithm SVM-SVOR-GenAge}
  \scalebox{1}
  {
\begin{tabular}{ll}
\hline
\hline
\MYnextline{l}{\textbf{Input:}\\ \\} & \MYnextline{l}{Training instances $X$, and labels $Y_{gender}$ and $Y_{age}$;\\ Parameters $\lambda_{1}$, $\lambda_{2}$, and $\lambda_{3}$.}\\
\textbf{Output:} & $w_{g}$, $b_{g}$, $w_{a}$, $b_{a}$.\\
\hline
\multicolumn{2}{l}{\MYnextline{l}{1.$\;$Initialize $w_{a}$, $b^{m}$, and $b^{f}$;\\
2.$\;$\textbf{for} $t=1, 2, ...,T_{max}$ \textbf{do}\\
3.$\quad \ \ $ Compute $w_{g}$ and $b_{g}$ based on \eqref{eq:SVM&SVM-SVOR-GenAge};\\
4.$\quad \ \ $ Compute $w_{a}$, $b^{m}$ and $b^{f}$ based on \eqref{eq:SVOR&SVM-SVOR-GenAge};\\
5.$\;$\textbf{end for}\\
6.$\;$Return $w_{g}$, $b_{g}$, $w_{a}$, $b^{m}$, and $b^{f}$.\\
}
} \\
\hline
\hline
\end{tabular}
   }
\end{table}

It is worth pointing out that in this work, since we emphasize on exploring the aging difference between the male and the female to age estimation, so we do not conduct kernelization for the proposed method.

\section{Experiment} \label{sec:experiment}
To evaluate the proposed method, in this section we conduct age estimation experiments on several human aging datasets.

\subsection{Datasets} \label{sec:dataset}
\begin{figure}[htdp!]
  \centering
  \includegraphics[width=0.95\linewidth]{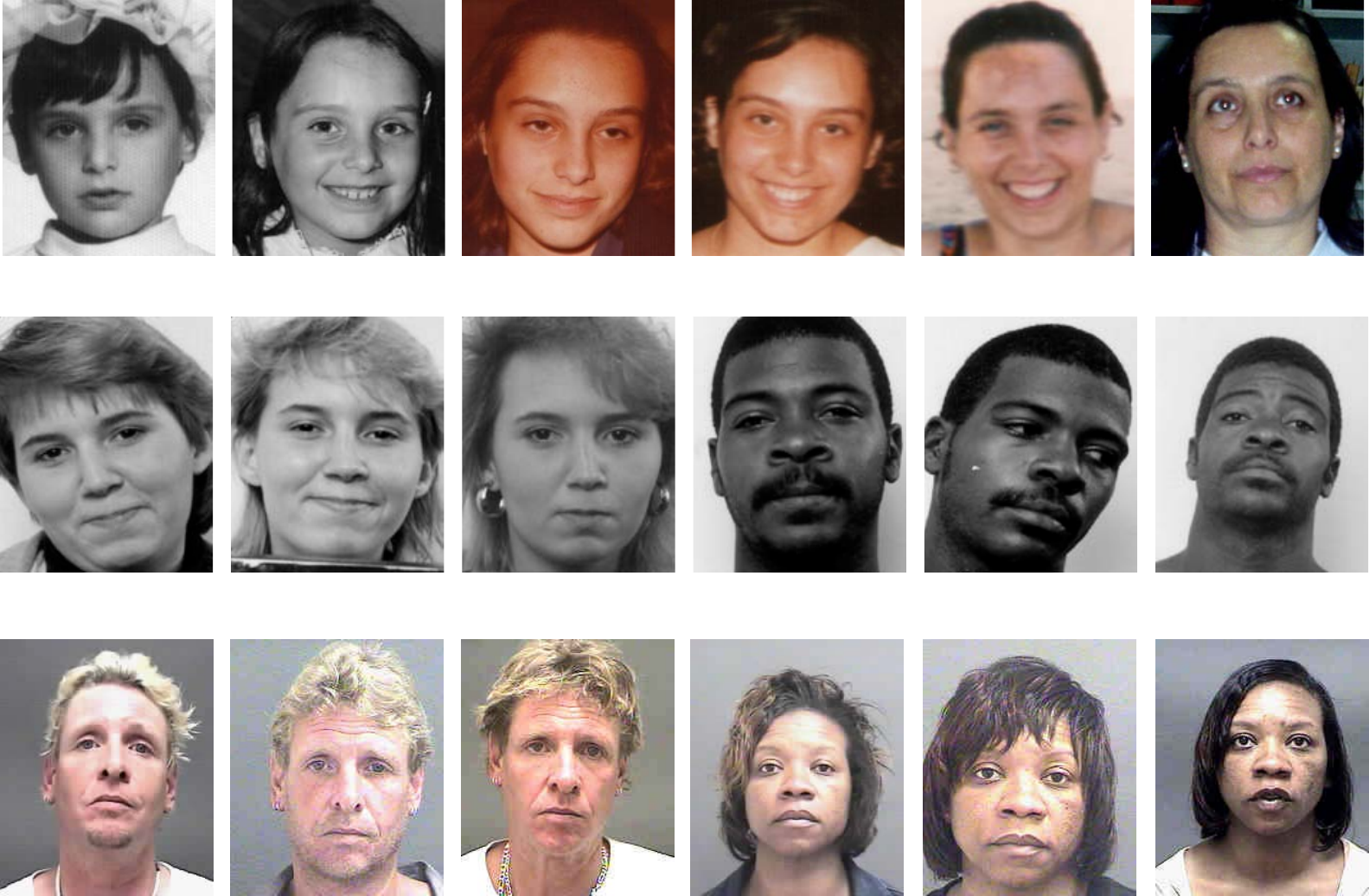}\\
  \caption{Image examples from FG-NET (1st row), Morph Album I (2nd row), and Morph Album II (3rd row).}\label{fig:dataset-examples}
\end{figure}

In the experiments, three well-known aging datasets, FG-NET, Morph Album I, and Morph Album II, were used. For FG-NET dataset, it consists of 1,002 facial images captured from 82 persons from 0 to 36 years old. For Morph Album I, there are about 1,690 facial images from about 631 persons from 16 to about 77 years old. Morph Album II is a relatively larger dataset with over 55,000 images and their ages range from 16 to about 77 years old. Some examples from the three datasets are shown in Figure \ref{fig:dataset-examples}.

\subsection{Experimental Setup} \label{sec:experimental setup}
In the experiments, the values of all the hyper-parameters\footnote{The optimal value of $\lambda_3$ involved in Eq. \eqref{eq:SVM-SVOR-GenAge} is suggested to tune with relatively large values, to emphasize the regularization term.} involved are set via 5-fold \emph{cross-validation}. And we set the $T_{max}$ in Table \ref{tab: algorithm SVM-SVOR-GenAge} to 2\footnote{In the experiments, we found that after about 2 iterations, the objective function value of Eq. \eqref{eq:SVM-SVOR-GenAge} converges.}. Besides, we uniformly adopt the Mean Absolute Errors (MAE) as the performance measure, where $MAE:= \frac{1}{N}\sum_{i=1}^{N}|\widehat{l}_i-l_i|$ with $l_i$ and $\widehat{l}_i$ denoting the ground-true and predicted age values, respectively. All the reported results are averaged over 10 runs, each with the same experimental setup. In order to adequately evaluate the proposed method, we introduce several methods for comparison as follows:
\begin{itemize}
  \item \textbf{PLS}: the method of \cite{guo2014framework}, which jointly regresses human gender and age as two concatenated output variables and is a state-of-the-art method.
  \item \textbf{DirectAgeEstimation}: the method of Eq. \eqref{eq:SVM-SVOR-GenAge} with $\lambda_3$ = 0 and $b^m$ = $b^f$, which directly performs age estimation without discriminating the gender.
  \item \textbf{2StepBasedAgeEstimation}: the method of Eq. \eqref{eq:SVM-SVOR-GenAge} with $\lambda_3$ = 0, which performs gender classification (GC) and then conducts male/female age estimation according to the GC result, and is a two-step method as \cite{guo2010human} and \cite{lakshmiprabha2011age}.
  \item \textbf{AgeEstimation-ST}: the method of Eq. \eqref{eq:SVM-SVOR-GenAge} with $b^m$ = $b^f$, which performs age estimation with preserving the semantic relations between the gender and age.
  \item \textbf{AgeEstimation-TT(Ours)}: the method of Eq. \eqref{eq:SVM-SVOR-GenAge}, which performs gender-aware age estimation in a unified framework with preserving not only the semantic relations between the gender and age, but also the aging discrepancy between the male and the female.
\end{itemize}

\subsection{Age Estimation With Race Variation} \label{sec:experiment with ethnicity variation}
\begin{figure*}[htdp!]
  \centering
  \subfigure[]{
    \label{fig:fgnet-total} 
    \includegraphics[width=0.315\linewidth, height=1.6in]{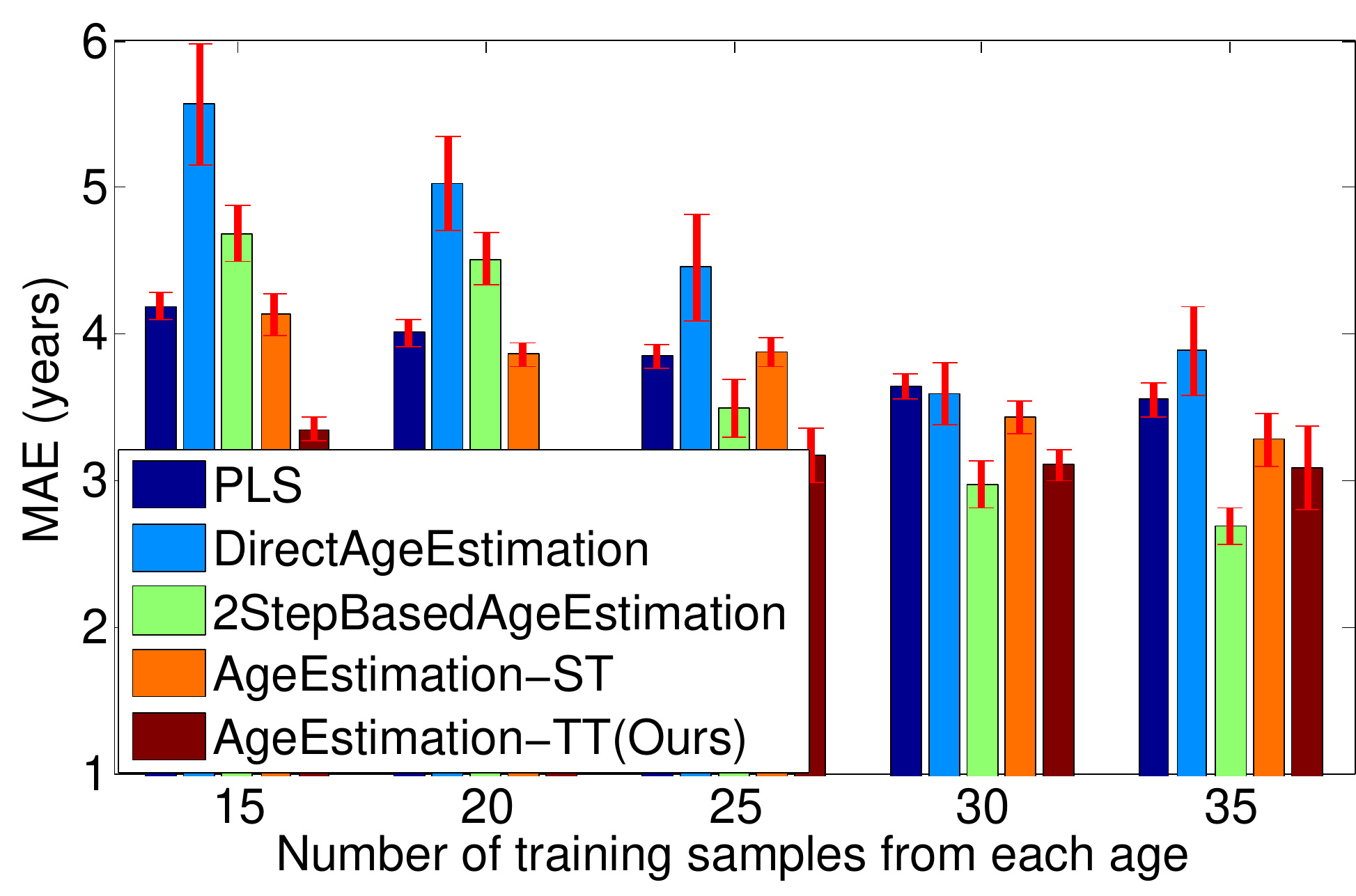}}
  \subfigure[]{
    \label{fig:fgnet-male} 
    \includegraphics[width=0.315\linewidth, height=1.6in]{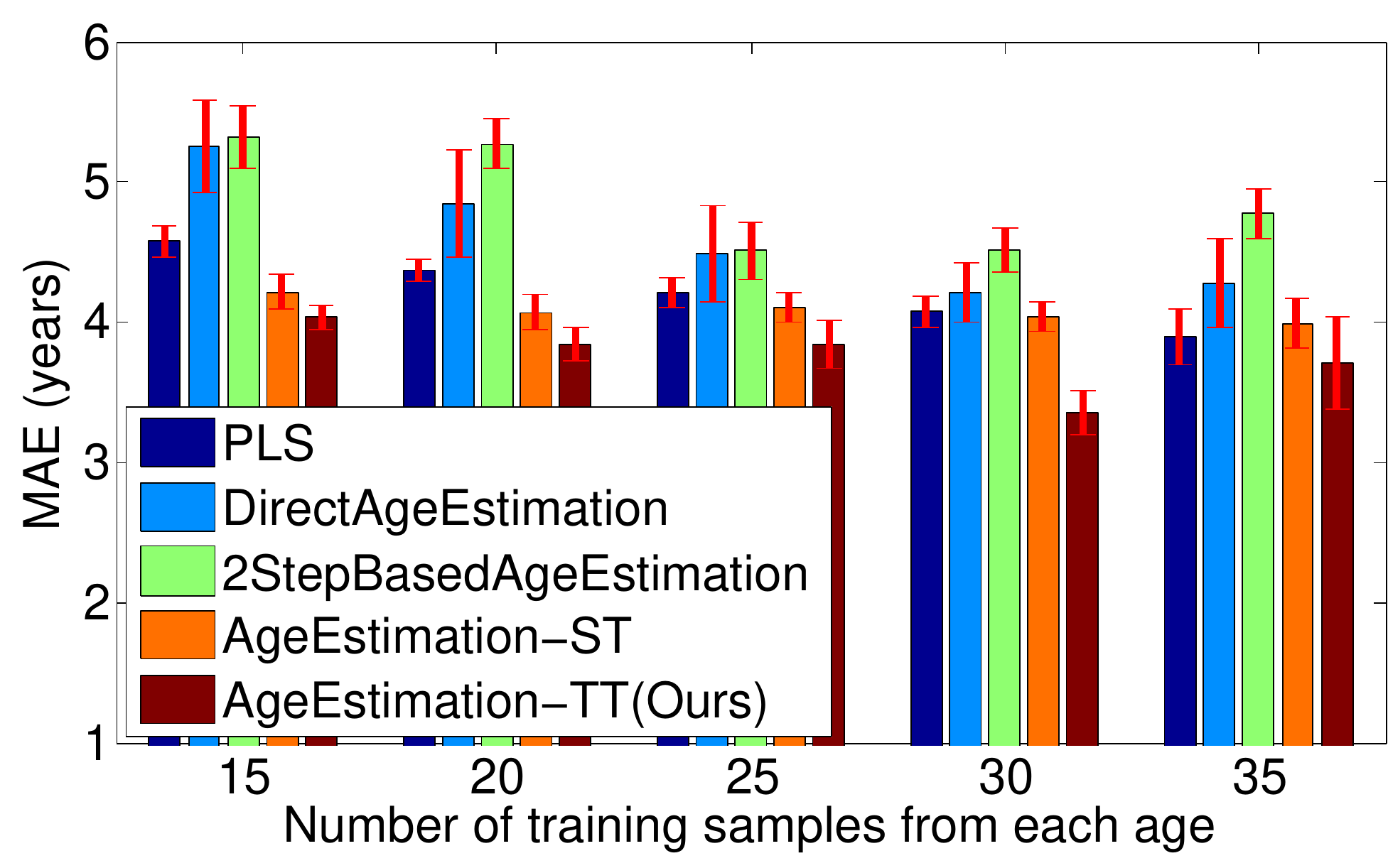}}
  \subfigure[]{
    \label{fig:fgnet-female} 
    \includegraphics[width=0.315\linewidth, height=1.6in]{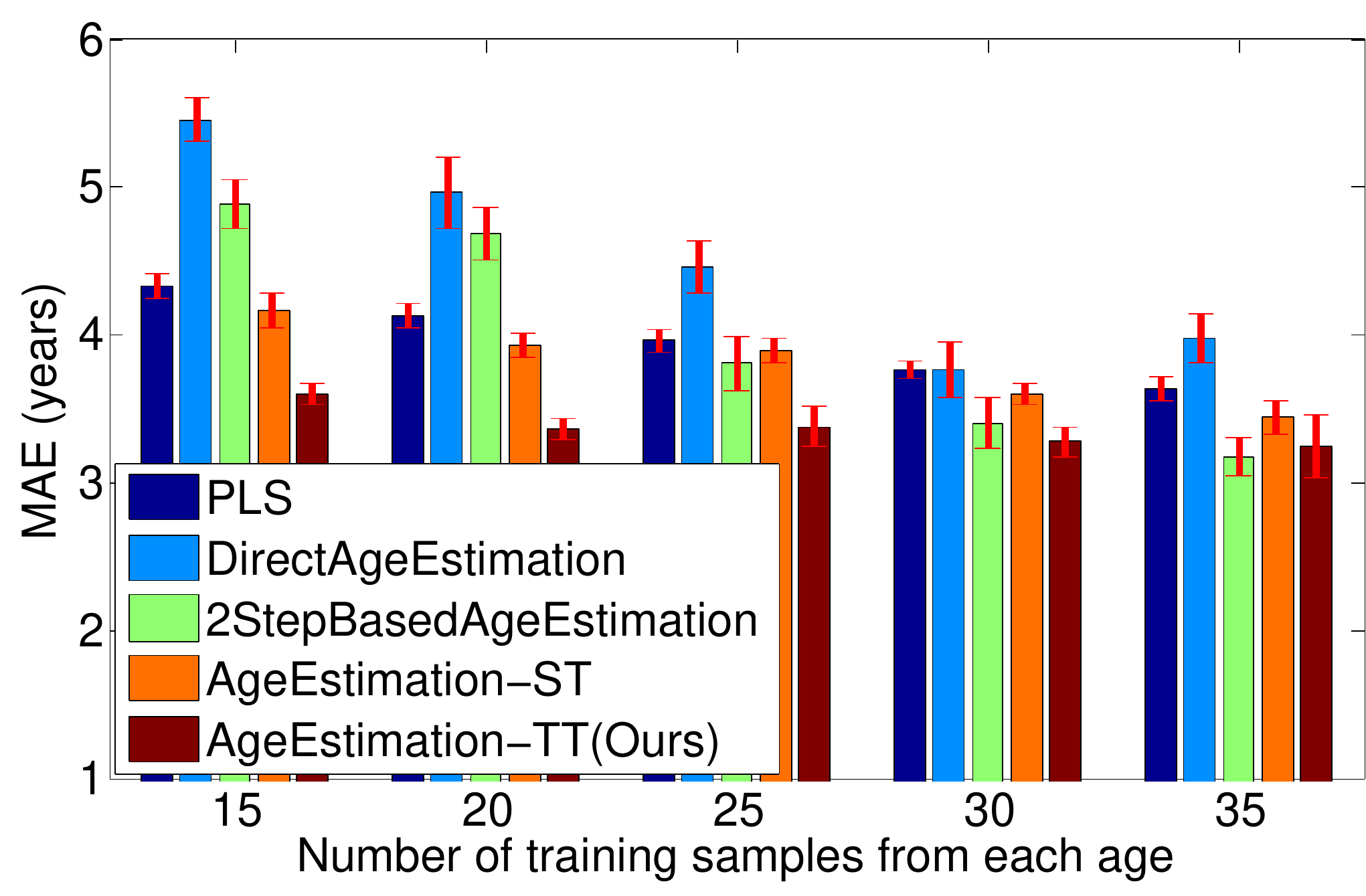}}
  \caption{Age estimation on FG-NET with MAE counted in terms of male (a), female (b), and their mixture (c).}
  \label{fig:fgnet-result} 
\end{figure*}

\begin{figure*}[htdp!]
  \centering
  \subfigure[]{
    \label{fig:fgnet-total} 
    \includegraphics[width=0.315\linewidth, height=1.6in]{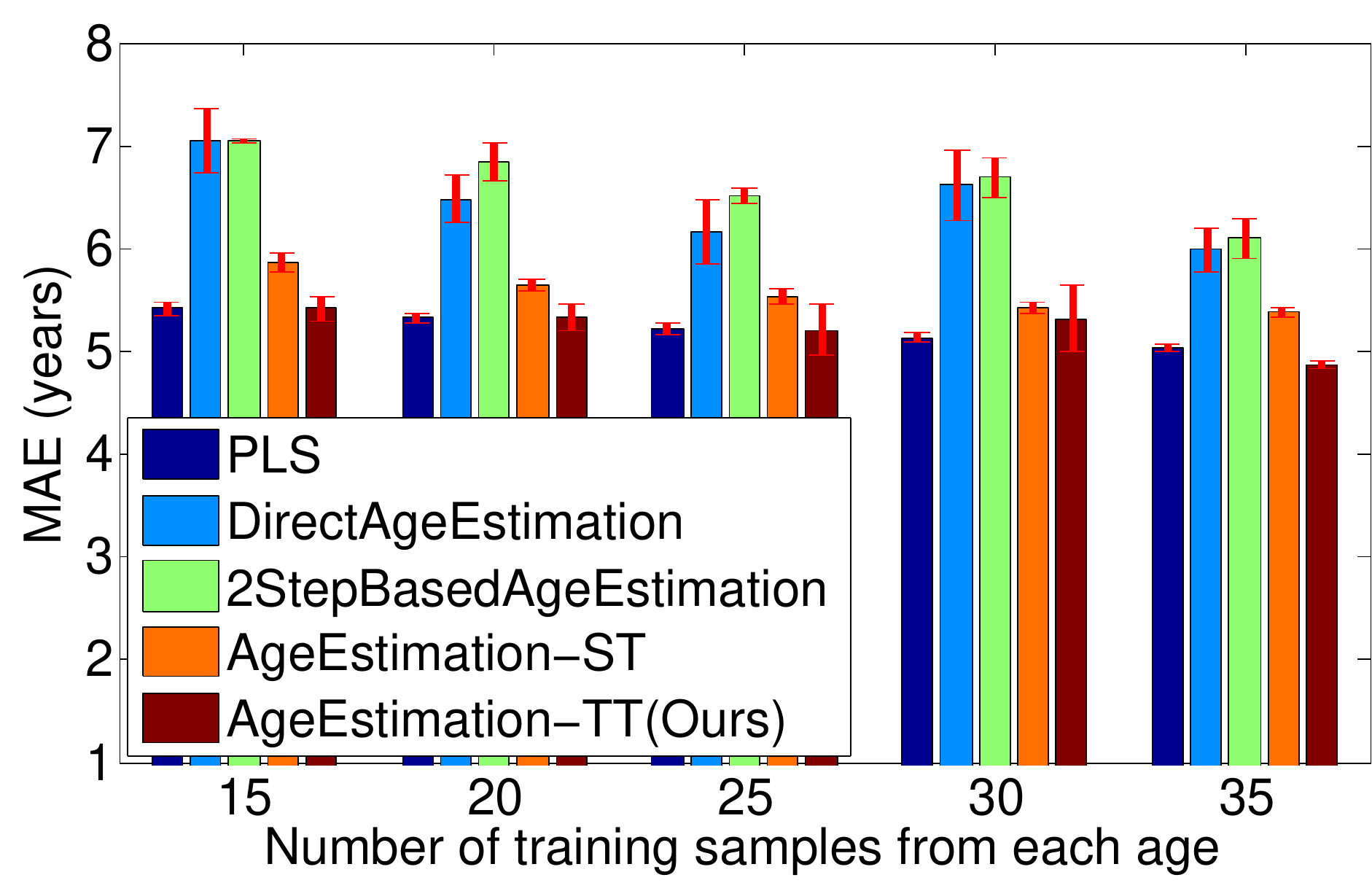}}
  \subfigure[]{
    \label{fig:fgnet-male} 
    \includegraphics[width=0.315\linewidth, height=1.6in]{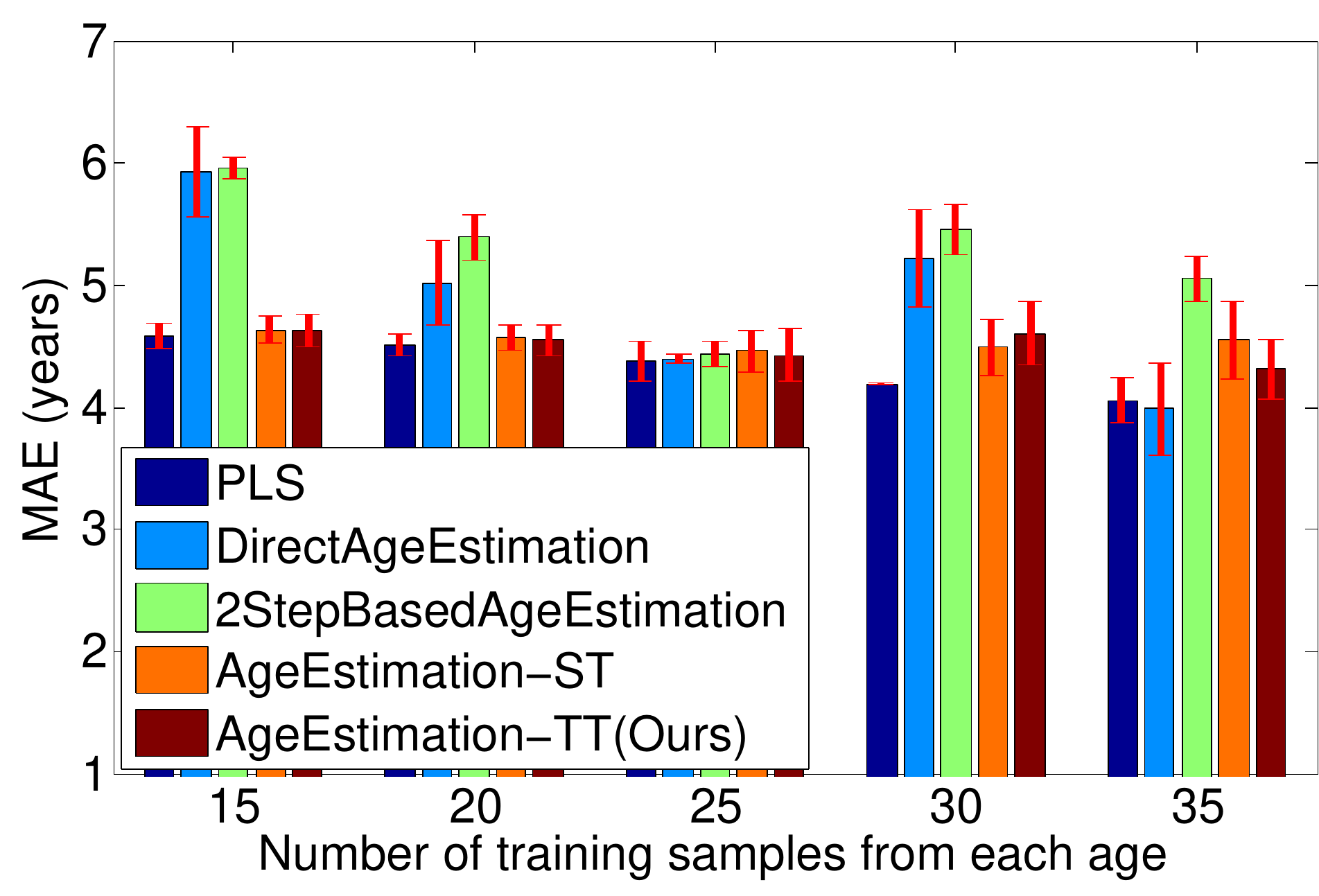}}
  \subfigure[]{
    \label{fig:fgnet-female} 
    \includegraphics[width=0.315\linewidth, height=1.6in]{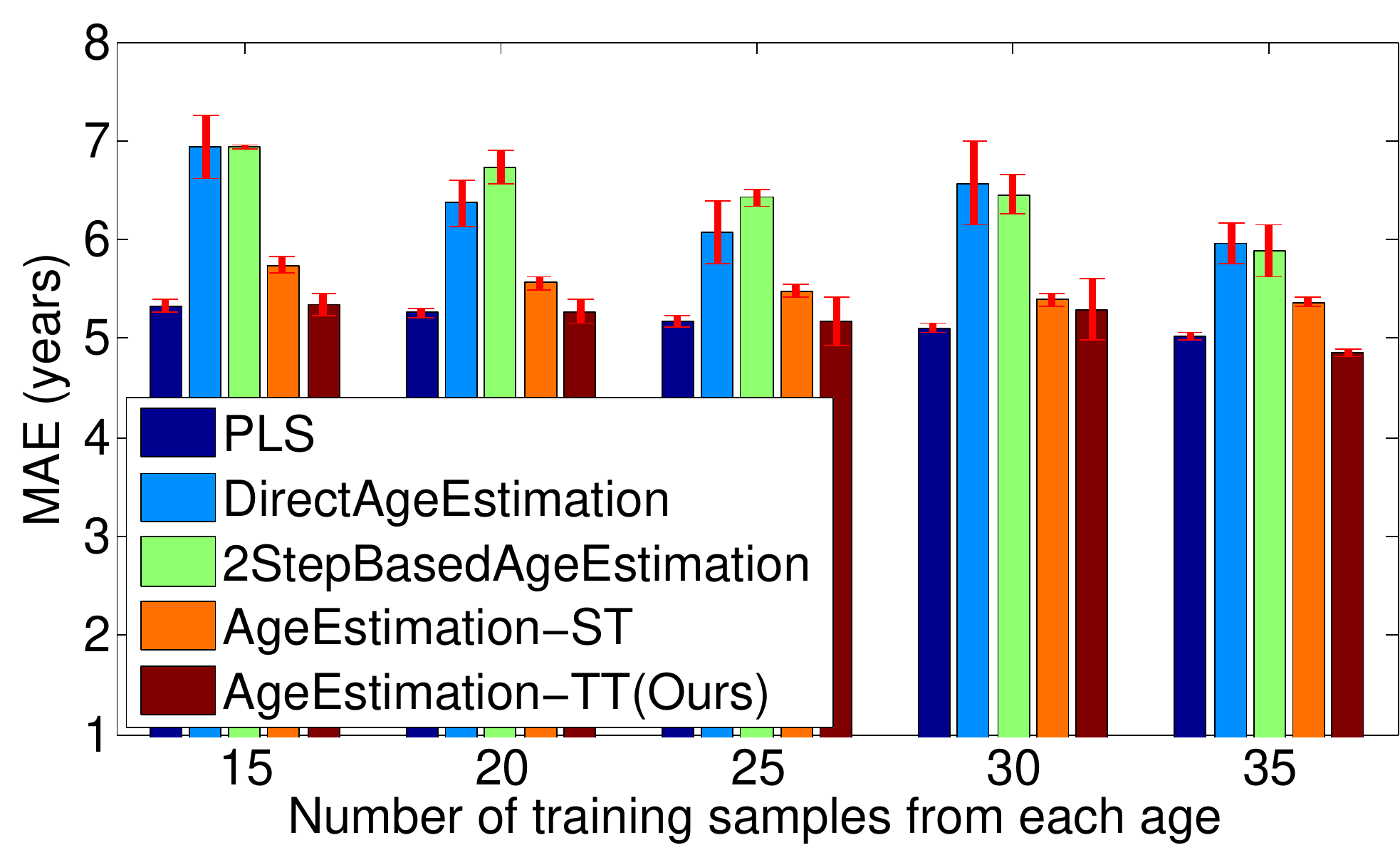}}
  \caption{Age estimation on Morph Album I with MAE counted in terms of male (a), female (b), and their mixture (c).}
  \label{fig:morph1-result} 
\end{figure*}

\begin{figure*}[htdp!]
  \centering
  \subfigure[]{
    \label{fig:fgnet-total} 
    \includegraphics[width=0.315\linewidth, height=1.6in]{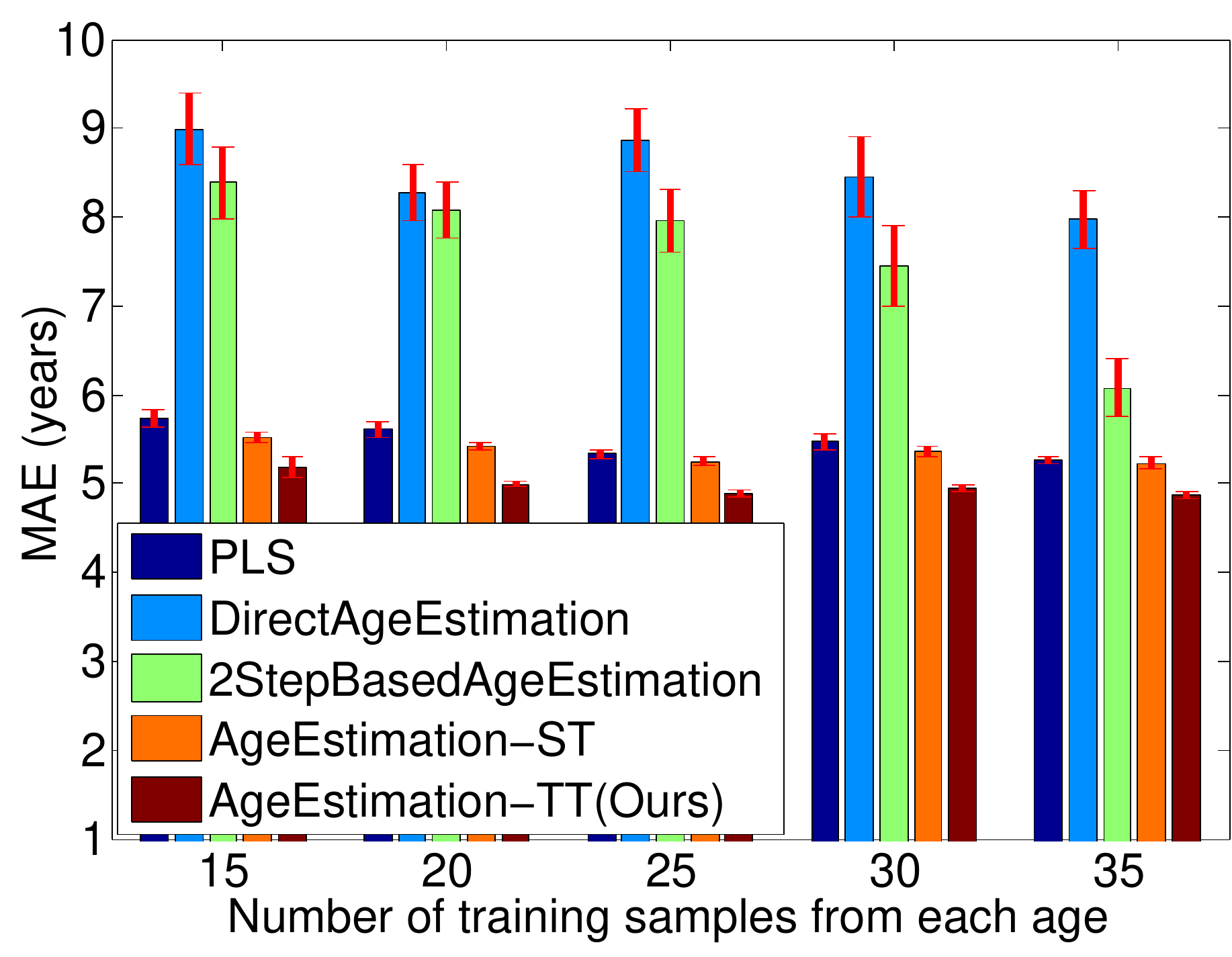}}
  \subfigure[]{
    \label{fig:fgnet-male} 
    \includegraphics[width=0.315\linewidth, height=1.6in]{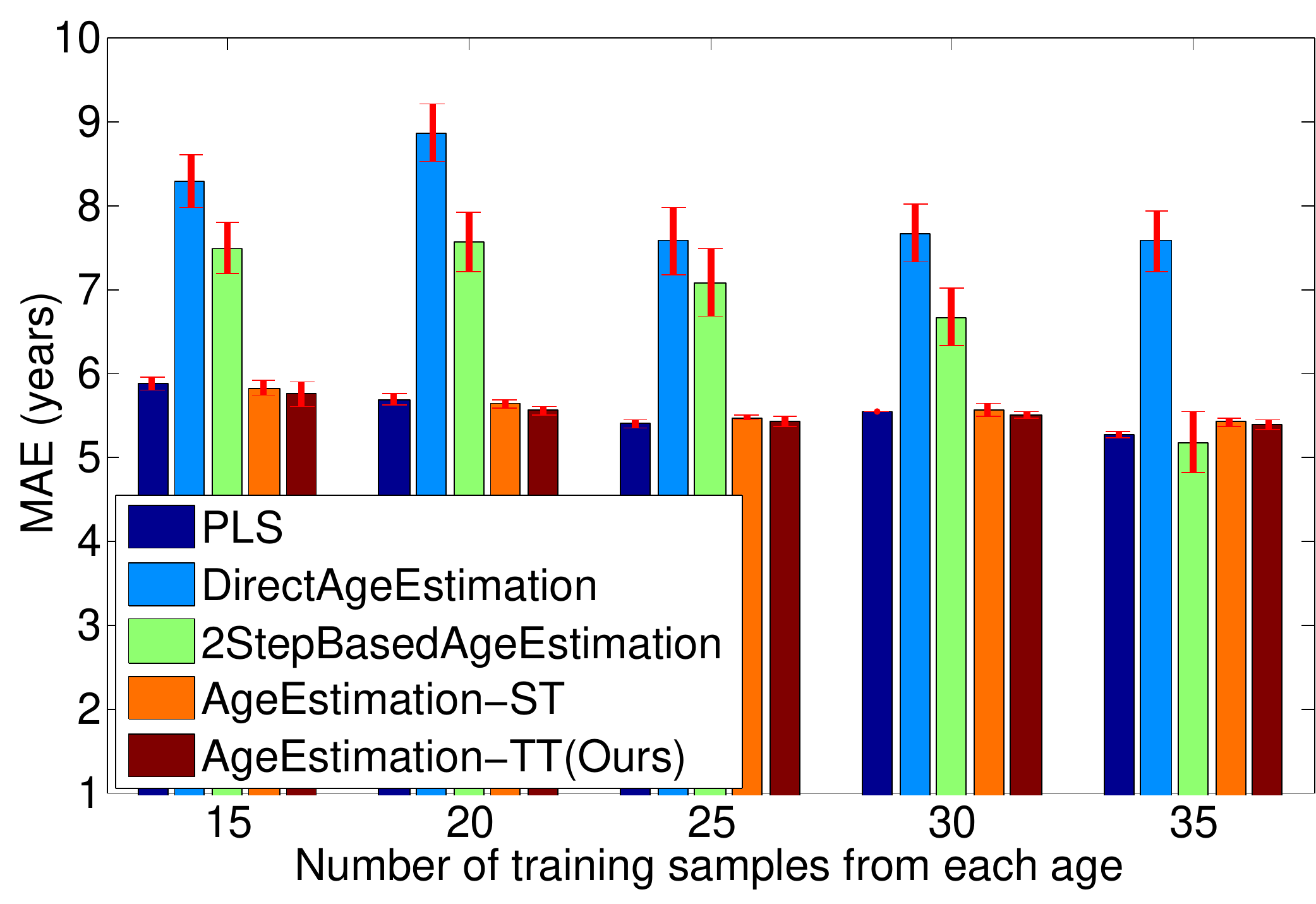}}
  \subfigure[]{
    \label{fig:fgnet-female} 
    \includegraphics[width=0.315\linewidth, height=1.6in]{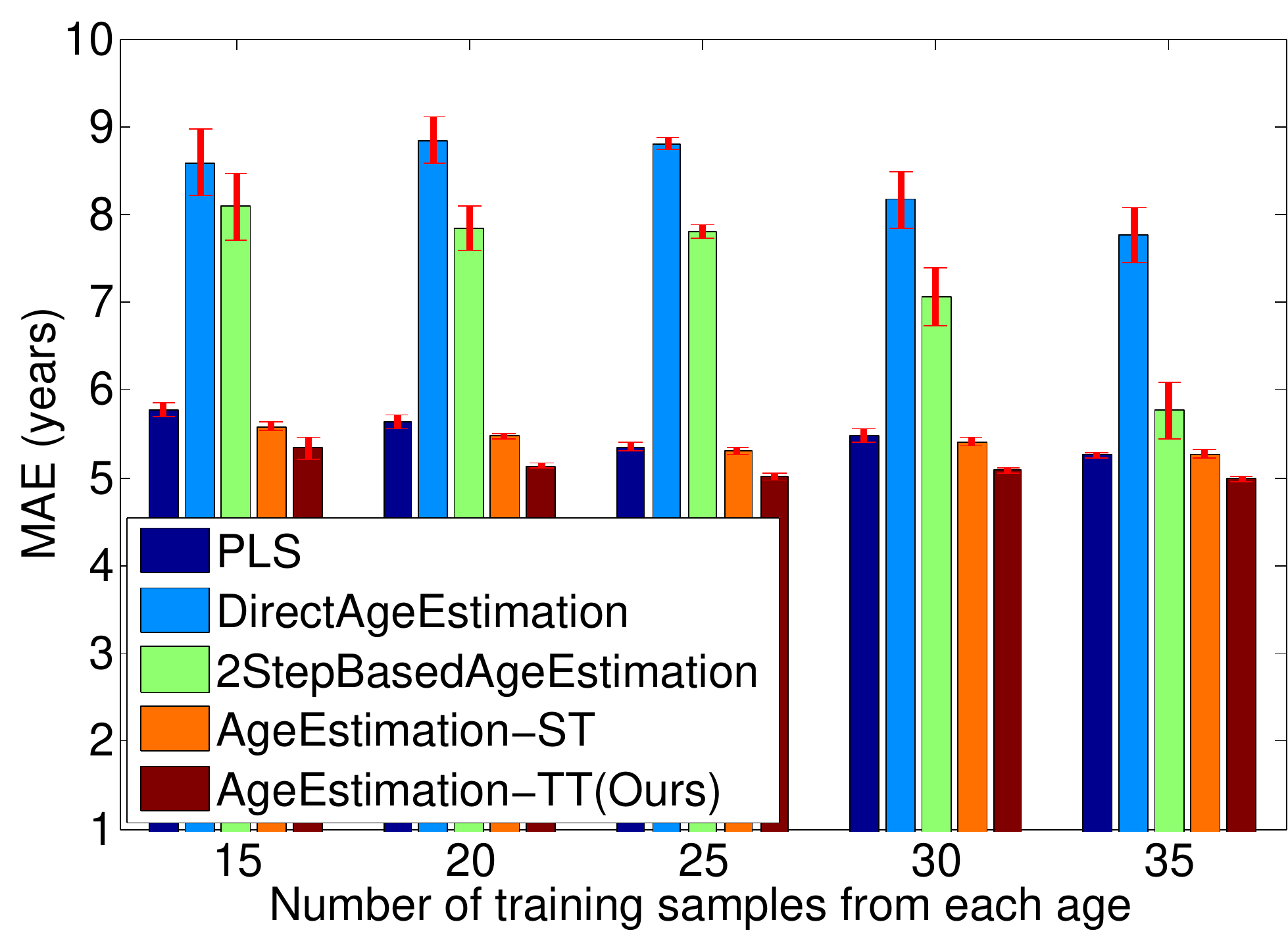}}
  \caption{Age estimation on Morph Album II with MAE counted in terms of male (a), female (b), and their mixture (c).}
  \label{fig:morph2-result} 
\end{figure*}

We first conduct age estimations with race variation involved. That is, the data used for training and testing are sampled across races, e.g., the white race, yellow race, and black race. We extract 200-dimensional AAM parameters from FG-NET and Morph Album I, and BIFs from Morph Album II as feature representations, respectively. We experiment with varying number of samples for training and the rest for testing\footnote{If the number of samples at some ages is less than required, all of their samples will be selected for training.}, and report the results in Figures \ref{fig:fgnet-result}, \ref{fig:morph1-result}, and \ref{fig:morph2-result}. From them, we can discover that
\begin{itemize}
  \item With increasing number of training data, the MAEs of all the methods are generally decreasing. It witnesses that increasing training data improve the generalization ability of an estimator.
  \item On FG-NET and Morph Album II, the MAEs of 2StepBasedAgeEstimation are generally lower than that of DirectAgeEstimation. It demonstrates that the male and the female are not aging synchronously, and that performing gender discriminating before age estimation can promote the latter's accuracy. By contrast, on Morph Album I, the MAEs of 2StepBasedAgeEstimation are dramatically higher than those of DirectAgeEstimation. It is because that the gender classification accuracy of 2StepBasedAgeEstimation on this dataset is somewhat low (less than 80\%), which deteriorates the consequent age estimation accuracy.
  \item In most cases, the MAEs of AgeEstimation-ST are greatly lower than those of 2StepBasedAgeEstimation. It shows that discriminating the semantic relation between human gender and age benefits to promote the age estimation accuracy.
  \item Generally, the proposed method, i.e., AgeEstimation-TT, yields the highest age estimation accuracy on the three aging datasets. It demonstrates the superiority of the proposed method in age estimation. More importantly, it also witnesses the soundness of incorporating the semantic relation between human gender and age, as well as the aging discrepancy between the male and the female in age estimation modeling.
  \item Comparing between Figures \ref{fig:fgnet-result}, \ref{fig:morph1-result} and \ref{fig:morph2-result}, it can be discovered that the MAEs on Morph Album I are basically level with those on Morph Album II, both of them are generally higher than that on FG-NET by about 1 years old. It demonstrates that FG-NET dataset is relatively easier to estimate than the other two datasets. By investigating the three datasets, we find that all the individuals of FG-NET are Caucasian, while the Morph Album I and II are sampled across white, yellow, and black races. It shows that race variation deteriorates the accuracy of age estimation.
\end{itemize}

\begin{figure}[htdp!]
  \centering
  \subfigure[FG-NET]{
    \label{fig:fgnet-t} 
    \includegraphics[width=0.935\linewidth, height=1.5in]{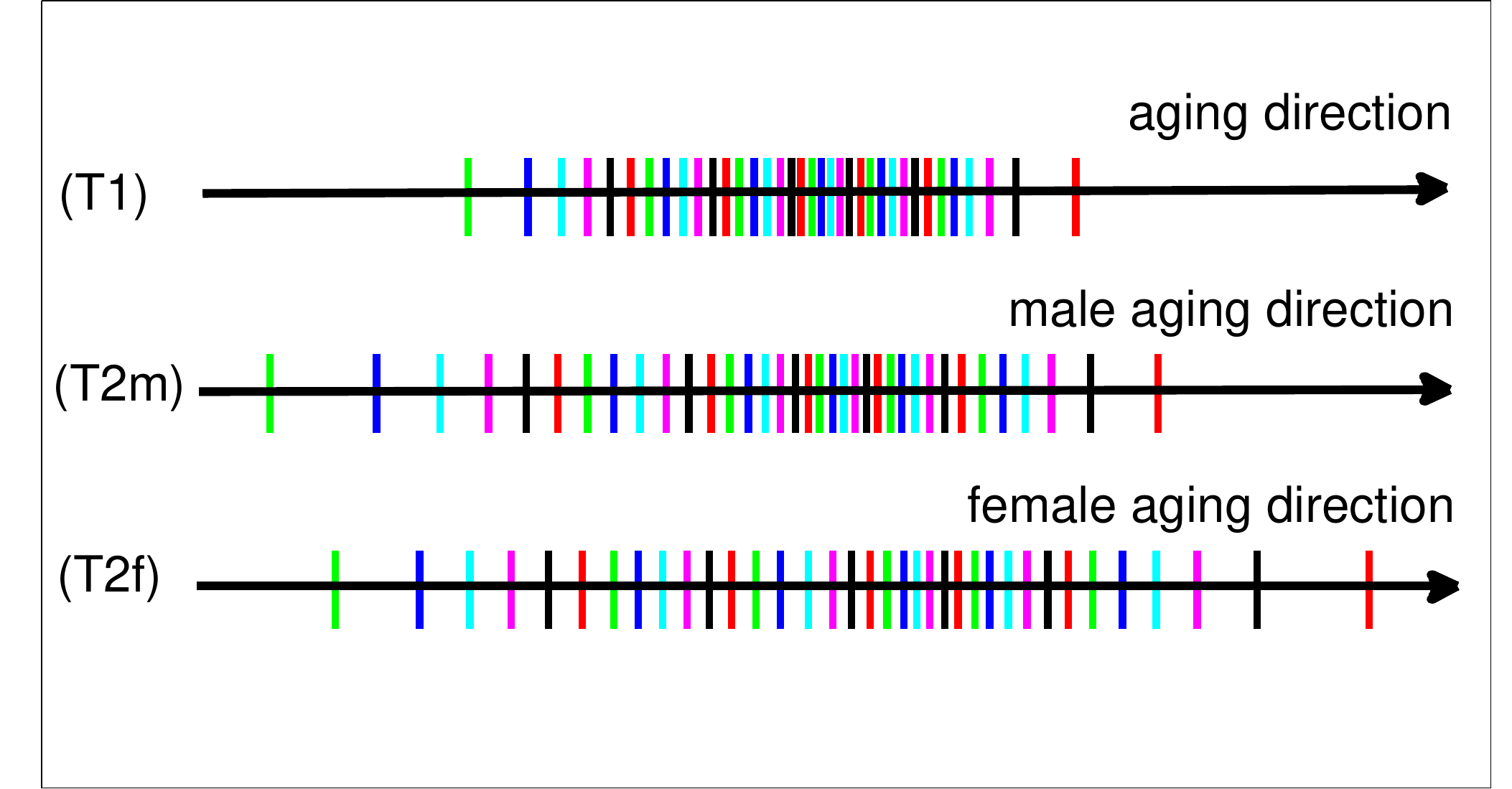}}
  \subfigure[Morph Album I]{
    \label{fig:morph1-t} 
    \includegraphics[width=0.95\linewidth, height=1.5in]{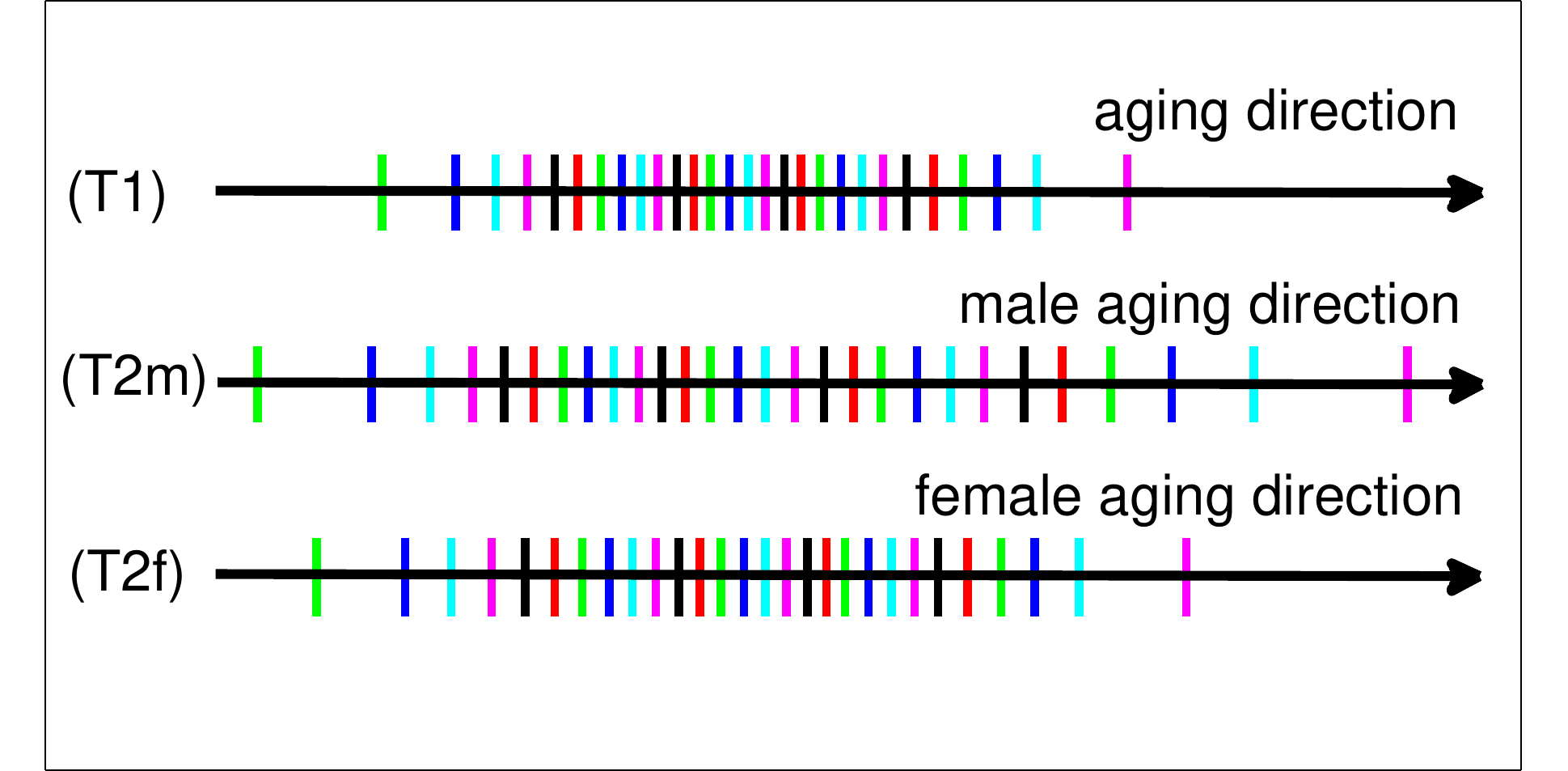}}
  \subfigure[Morph Album II]{
    \label{fig:morph2-t} 
    \includegraphics[width=0.95\linewidth, height=1.5in]{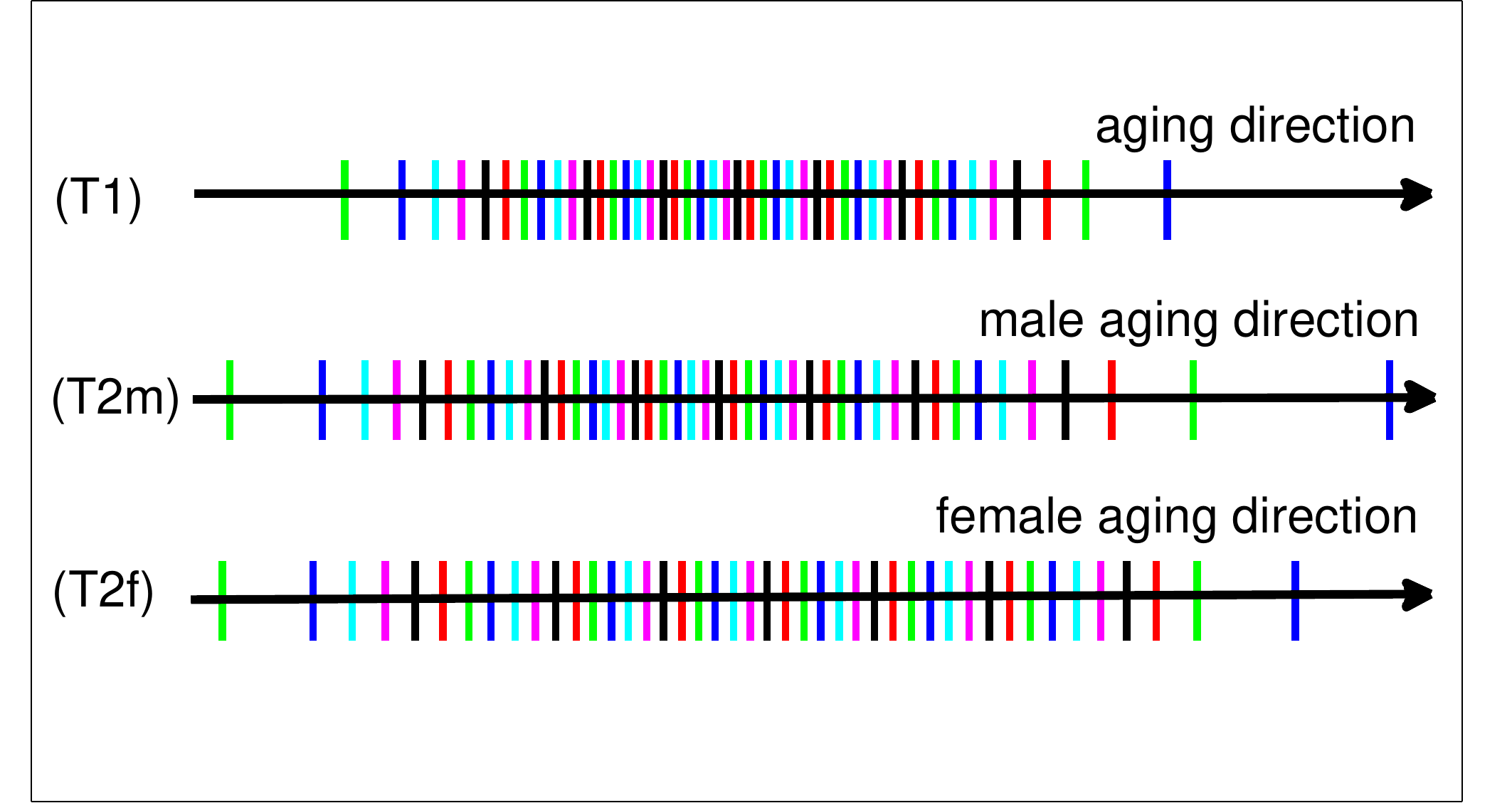}}
  \caption{Aging thresholds for the male (T2m) and female (T2f), respectively, learned by our method, and for their mixture (T1) learned by AgeEstimation-ST. The horizontal axis indicates the aging direction from young to aged (i.e., 0 to 36 years old on FG-NET (a), 16 to 44 years old on Morph Album I (b), and 16 to 60 years old on Morph Album II (c)), and each of the color bars represents one aging threshold between two neighboring ages.}
  \label{fig:threshold-plot} 
\end{figure}

Besides the above quantitative evaluations, we also visually explore the aging discrepancy between the male and the female. To this end, we compare the aging thresholds learned by AgeEstimation-ST and our proposed method (i.e., AgeEstimation-TT) in Figure \ref{fig:threshold-plot}, from which we can discover that
\begin{itemize}
  \item The set of aging thresholds (see (T1)), corresponding to the $b$ learned by AgeEstimation-ST for the male and the female together, are distributed more compact than those for the male (see (T2m)) and female (see (T2f)), respectively, corresponding to the $b^m$  and $b^f$ in Eq. \eqref{eq:SVM-SVOR-GenAge} learned by our method. Intuitively, compared with distribution-compact aging thresholds, more scattered are preferable because it reduces the difficulty of distinguishing the ages, which coincides with the results (see Figures \ref{fig:fgnet-result}, \ref{fig:morph1-result}, and \ref{fig:morph2-result}) that the age estimation accuracy by our method is higher than that by AgeEstimation-ST.
  \item The aging thresholds distributions of the male (see (T2m)) are significantly discrepant from those of the female (see (T2f)). It visually verifies the soundness of performing age estimation with taking into account the prior knowledge that the male and the female are aging quite different.
\end{itemize}

\subsection{Age Estimation Without Race Variation} \label{sec:experiment without ethnicity variation}
We also conduct age estimation without race variation to further explore the aging difference between the male and the female. To be specific, we select white race and black race\footnote{Apart from the white and black races, samples of other races in Morph Album II are too few, so we do not take them for experiment.} samples from Morph Album II for experiment, respectively. And we categorize the samples into five age groups (i.e., 0-19, 20-29, 30-39, 40-49, and 50+ years) for white race and black race, respectively. By this setting, the race variation is factored out. We conduct experiments using the generated data and report the results in Tables \ref{tab:white-age-group} and \ref{tab:black-age-group}. From them, we discover that:
\begin{itemize}
  \item Generally, the MAEs of age group estimation yielded by all the competing methods are decreasing with increasing samples, which once again witnesses that relatively more training samples benefit the training of an estimator and thus improve its generalization ability.
  \item MAEs of 2StepBasedAgeEstimation and AgeEstimation-ST both are lower than those of DirectAgeEstimation, and MAEs of 2StepBasedAgeEstimation are greatly lower than that of AgeEstimation-ST and are even lower than that of PLS. It demonstrates that although the semantic relation between the gender and age as well as the aging discrepancy between the male and the female both affect the performance of age estimation, the latter plays a bigger role when relatively a few age classes are involved.
  \item Overall, the proposed method, i.e., AgeEstimation-TT, yields the lowest MAEs in all cases. Surprisingly, the MAEs of the proposed method are even lower by over 10\% than those of PLS. Even better, the MAEs by our method with number of training samples from each age group equal to 50 are lower than those of PLS with number of each age group training samples equal to 150. It shows that explicitly incorporating not only the semantic relation between human gender and age, but also the gender discrepancy in aging in age estimation is reasonable and desirable.
  \item Comparing between Tables \ref{tab:white-age-group} and \ref{tab:black-age-group}, we can find that MAEs on white race are significantly lower than that on black race, which witnesses that the Caucasian's ages are relatively easier to discriminate than the Melanoderm from their facial appearances.
\end{itemize}
\begin{table*}[htbp!]
\centering
\caption{Age group estimation (MAE$\pm$STD) on white race.}\label{tab:white-age-group}
  \scalebox{0.95}
  {
  \begin{tabular}{llllll}
\hline
\multicolumn{1}{|c|}{\MYnextline{c}{\# training\\samples from\\each age group}} & \multicolumn{1}{c}{PLS} & \multicolumn{1}{c}{\MYnextline{c}{Direct-\\AgeEstimation}} & \multicolumn{1}{c}{\MYnextline{c}{2StepBased-\\AgeEstimation}} & \multicolumn{1}{c}{\MYnextline{c}{AgeEstimation-\\ST}} & \multicolumn{1}{c|}{\MYnextline{c}{AgeEstimation-\\TT(Ours)}} \\
\hline
\hline
\multicolumn{1}{|c|}{\emph{50}} & \multicolumn{1}{c}{\MYnextline{c}{0.99$\pm$0.02\\1.03$\pm$0.02\\0.88$\pm$0.04}} & \multicolumn{1}{c}{\MYnextline{c}{1.15$\pm$0.10\\1.16$\pm$0.09\\1.09$\pm$0.12}} &
\multicolumn{1}{c}{\MYnextline{c}{1.02$\pm$0.03\\1.05$\pm$0.02\\0.88$\pm$0.08}} &
\multicolumn{1}{c}{\MYnextline{c}{1.13$\pm$0.02\\1.15$\pm$0.02\\1.08$\pm$0.13}} & \multicolumn{1}{c|}{\MYnextline{c}{\textbf{0.88$\pm$0.03}\\\textbf{0.93$\pm$0.02}\\\textbf{0.70$\pm$0.04}}} \\
\hline
\multicolumn{1}{|c|}{\emph{100}} & \multicolumn{1}{c}{\MYnextline{c}{0.94$\pm$0.01\\0.99$\pm$0.01\\0.78$\pm$0.02}} & \multicolumn{1}{c}{\MYnextline{c}{1.14$\pm$0.09\\1.19$\pm$0.02\\1.08$\pm$0.09}} &
\multicolumn{1}{c}{\MYnextline{c}{0.97$\pm$0.04\\1.09$\pm$0.05\\0.82$\pm$0.07}} &
\multicolumn{1}{c}{\MYnextline{c}{1.06$\pm$0.05\\1.12$\pm$0.08\\0.92$\pm$0.09}} & \multicolumn{1}{c|}{\MYnextline{c}{\textbf{0.86$\pm$0.02}\\\textbf{0.91$\pm$0.01}\\\textbf{0.64$\pm$0.05}}} \\
\hline
\multicolumn{1}{|c|}{\emph{150}} & \multicolumn{1}{c}{\MYnextline{c}{0.91$\pm$0.01\\0.97$\pm$0.01\\0.70$\pm$0.02}} & \multicolumn{1}{c}{\MYnextline{c}{1.30$\pm$0.20\\1.32$\pm$0.32\\1.13$\pm$0.41}} &
\multicolumn{1}{c}{\MYnextline{c}{0.89$\pm$0.06\\0.94$\pm$0.05\\0.61$\pm$0.08}} &
\multicolumn{1}{c}{\MYnextline{c}{1.04$\pm$0.17\\1.08$\pm$0.15\\0.88$\pm$0.27}} & \multicolumn{1}{c|}{\MYnextline{c}{\textbf{0.84$\pm$0.01}\\\textbf{0.91$\pm$0.01}\\\textbf{0.58$\pm$0.03}}} \\
\hline
\end{tabular}
  }
    \begin{tablenotes}
  \footnotesize
  \item[1] \textbf{Note}: In each of the three training number cases, the 2nd, 3rd, and 1st row results are counted in terms of the male, female and their mixture, respectively.
  \end{tablenotes}
\end{table*}
\begin{table*}[htbp!]
\centering
\caption{Age group estimation (MAE$\pm$STD) on black race.}\label{tab:black-age-group}
  \scalebox{0.95}
  {
  \begin{tabular}{llllll}
\hline
\multicolumn{1}{|c|}{\MYnextline{c}{\# training\\samples from\\each age group}} & \multicolumn{1}{c}{PLS} & \multicolumn{1}{c}{\MYnextline{c}{Direct-\\AgeEstimation}} & \multicolumn{1}{c}{\MYnextline{c}{2StepBased-\\AgeEstimation}} & \multicolumn{1}{c}{\MYnextline{c}{AgeEstimation-\\ST}} & \multicolumn{1}{c|}{\MYnextline{c}{AgeEstimation-\\TT(Ours)}} \\
\hline
\hline
\multicolumn{1}{|c|}{\emph{50}} & \multicolumn{1}{c}{\MYnextline{c}{1.08$\pm$0.02\\1.09$\pm$0.02\\0.98$\pm$0.02}} & \multicolumn{1}{c}{\MYnextline{c}{1.37$\pm$0.09\\1.38$\pm$0.10\\1.23$\pm$0.08}} &
\multicolumn{1}{c}{\MYnextline{c}{1.10$\pm$0.09\\1.15$\pm$0.11\\0.95$\pm$0.04}} &
\multicolumn{1}{c}{\MYnextline{c}{1.17$\pm$0.12\\1.21$\pm$0.08\\1.06$\pm$0.03}} & \multicolumn{1}{c|}{\MYnextline{c}{\textbf{0.98$\pm$0.06}\\\textbf{1.00$\pm$0.06}\\\textbf{0.88$\pm$0.06}}} \\
\hline
\multicolumn{1}{|c|}{\emph{100}} & \multicolumn{1}{c}{\MYnextline{c}{1.03$\pm$0.01\\1.05$\pm$0.01\\0.92$\pm$0.01}} & \multicolumn{1}{c}{\MYnextline{c}{1.31$\pm$0.08\\1.36$\pm$0.09\\1.21$\pm$0.12}} &
\multicolumn{1}{c}{\MYnextline{c}{1.02$\pm$0.07\\1.09$\pm$0.09\\0.87$\pm$0.04}} &
\multicolumn{1}{c}{\MYnextline{c}{1.12$\pm$0.09\\1.15$\pm$0.07\\1.04$\pm$0.09}} & \multicolumn{1}{c|}{\MYnextline{c}{\textbf{0.96$\pm$0.02}\\\textbf{0.98$\pm$0.02}\\\textbf{0.84$\pm$0.03}}} \\
\hline
\multicolumn{1}{|c|}{\emph{150}} & \multicolumn{1}{c}{\MYnextline{c}{1.01$\pm$0.01\\1.03$\pm$0.01\\0.89$\pm$0.01}} & \multicolumn{1}{c}{\MYnextline{c}{1.25$\pm$0.14\\1.28$\pm$0.12\\1.09$\pm$0.11}} &
\multicolumn{1}{c}{\MYnextline{c}{0.98$\pm$0.04\\1.05$\pm$0.07\\0.87$\pm$0.03}} &
\multicolumn{1}{c}{\MYnextline{c}{1.06$\pm$0.09\\1.10$\pm$0.10\\0.93$\pm$0.04}} & \multicolumn{1}{c|}{\MYnextline{c}{\textbf{0.95$\pm$0.02}\\\textbf{0.97$\pm$0.02}\\\textbf{0.82$\pm$0.02}}} \\
\hline
\end{tabular}
  }
    \begin{tablenotes}
  \footnotesize
  \item[1] \textbf{Note}: In each of the three training number cases, the 2nd, 3rd, and 1st row results are counted in terms of the male, female and their mixture, respectively.
  \end{tablenotes}
\end{table*}

\begin{figure}[htdp!]
  \centering
  \subfigure[On white race]{
    \label{fig:white-group-t} 
    \includegraphics[width=0.90\linewidth, height=1.3in]{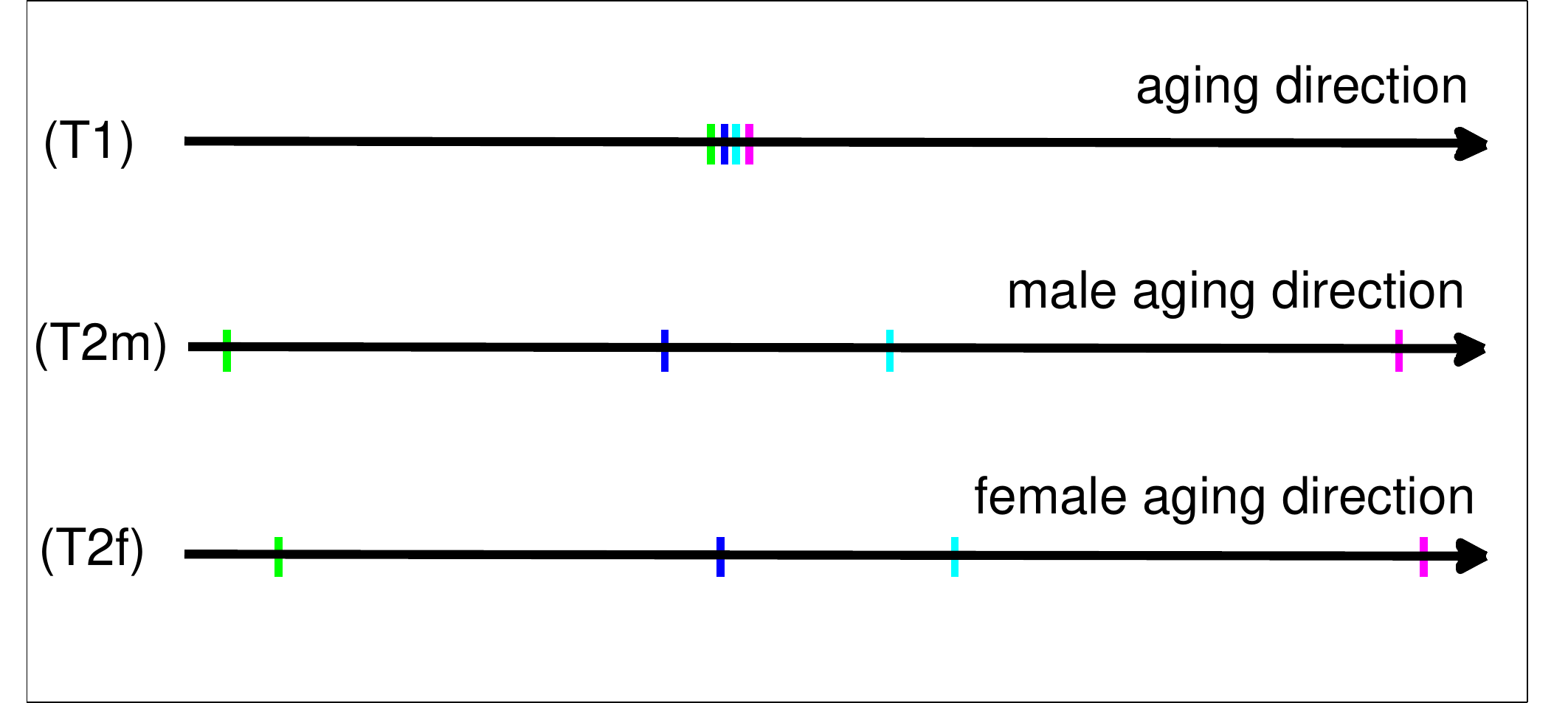}}
  \subfigure[On black race]{
    \label{fig:black-group-t} 
    \includegraphics[width=0.90\linewidth, height=1.3in]{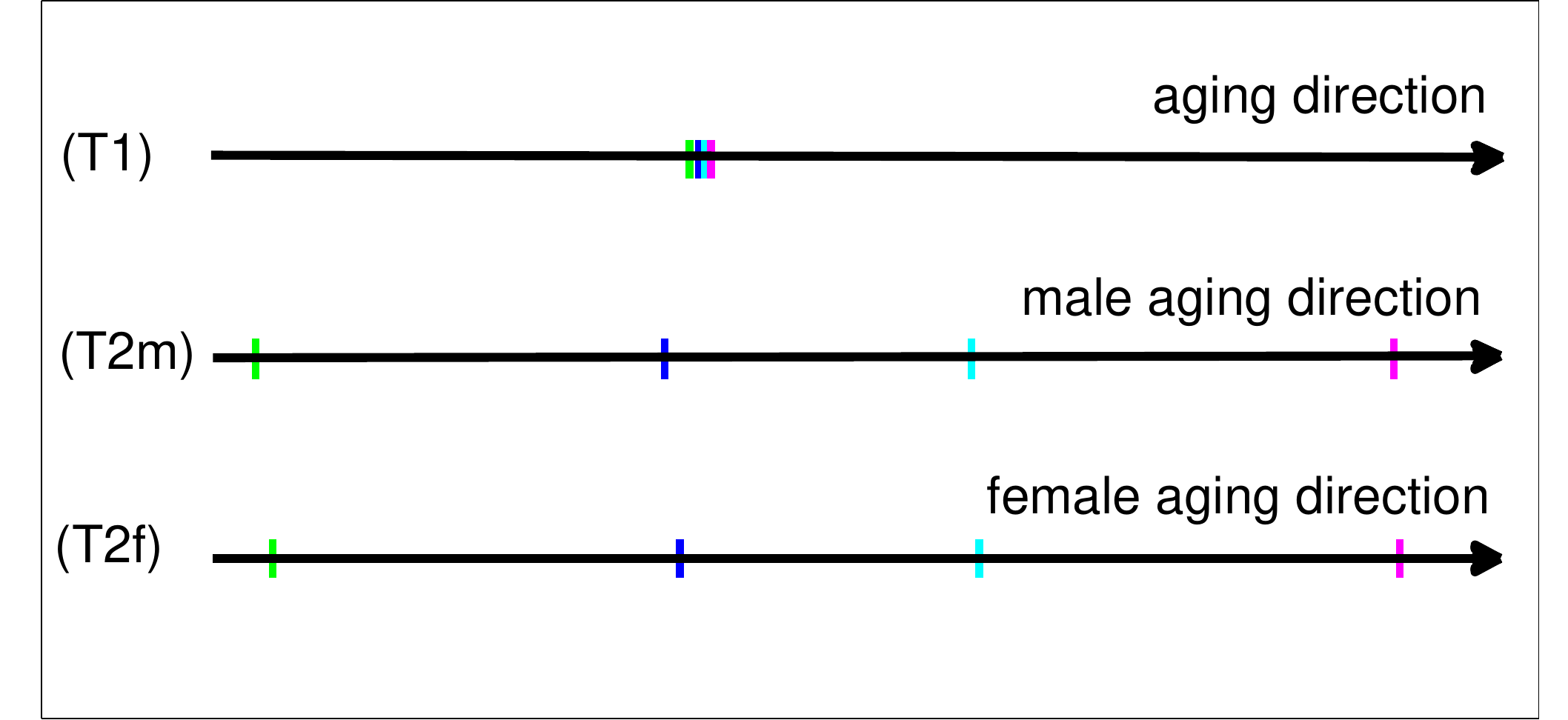}}
  \caption{Aging thresholds between five age groups learned by our method ((T2m) and (T2f)) and by the AgeEstimation-ST
 (T1). Each of the color bars represents an aging threshold between neighboring two of the five age groups (0-19, 20-29, 30-39, 40-49, and 50+ years old).}
  \label{fig:ethnicity-group-threshold} 
\end{figure}

Besides, we also display the learned aging thresholds on the white and black races in Figures \ref{fig:white-group-t} and \ref{fig:black-group-t}, respectively. From them, we can find that the aging difference between the male and the female on the white race is more obvious than the black race (see Figure \ref{fig:ethnicity-group-threshold}), which coincides with the observation that it is more difficult to tell a Melanoderm's age than the Caucasian.

\subsection{Reasonableness of Near-Orthogonality Penalty Across Human Gender and Age Semantic Spaces} \label{sec:angle analysis}
In this section, we verify the near-orthogonality between the gender and age semantic spaces. To this end, denote as $\theta$ the intersection angle between human gender discriminant direction $w_g$ and aging direction $w_a$. Then we conduct experiments to analyze $\theta$ by choosing 25 training samples from each age on FG-NET, Morph Album I and Album II, respectively. And the results are summarized in Table \ref{tab:angle-between-wg-and-wa}.
\begin{table}[htbp!]
\centering
\caption{Comparison of intersection angle $\theta$ between the gender discriminant direction $w_g$ and aging direction $w_a$.}\label{tab:angle-between-wg-and-wa}
  \scalebox{0.95}
  {
\begin{tabular}{llll}
\hline
\multicolumn{1}{|c|}{--} & \multicolumn{1}{c}{PLS} & \multicolumn{1}{c}{\MYnextline{c}{2StepBased-\\AgeEstimation}} & \multicolumn{1}{c|}{Ours} \\
\hline
\hline
\multicolumn{1}{|c|}{FG-NET} & \multicolumn{1}{c}{$134^o$} & \multicolumn{1}{c}{$97^o$} & \multicolumn{1}{c|}{$91^o$} \\
\multicolumn{1}{|c|}{Morph Album I} & \multicolumn{1}{c}{$99^o$} & \multicolumn{1}{c}{$70^o$} & \multicolumn{1}{c|}{$90^o$} \\
\multicolumn{1}{|c|}{Morph Album II} & \multicolumn{1}{c}{$120^o$} & \multicolumn{1}{c}{$79^o$} & \multicolumn{1}{c|}{$90^o$} \\
\hline
\end{tabular}
  }
  \begin{tablenotes}
  \footnotesize
  \item[1] \textbf{Note}: For PLS, human gender discriminant direction $w_g$ and aging direction $w_a$ refer to the corresponding regression weight vectors.
  \end{tablenotes}
\end{table}

From the results we find that generally, the $\theta$ learned by our method lies near to $90^o$ (coincides with the illustration in Figure \ref{fig:flowchart-of-GenAge}), while the $\theta$'s learned via the PLS and 2StepBasedAgeEstimation lie far from $90^o$.
Recalling the performance superiority of our method in previous experiments, we can see that Table \ref{tab:angle-between-wg-and-wa} witnesses the reasonableness of performing gender-aware age estimation in a estimation space nearly-orthogonal to gender semantic space, and they also provide an intuitive explanation of why we should set the $\lambda_3$ in Eq. \eqref{eq:SVM-SVOR-GenAge} with a relatively large value.

\section{Conclusion} \label{sec:conclusions}
In this paper, motivated by the fact that the male and the female are aging quite differently, we proposed a unified framework for gender-aware age estimation, in which the whole gender space is separated into male and female gender subspaces while gender-specific age estimation is performed in the gender subspaces. Then, we exemplified the framework to perform gender-aware age estimation, by which not only the semantic relationship between the gender and the age, but also the aging discrepancy between the male and the female are utilized in age estimation. Finally, experimental results demonstrated not only the superiority of our method in age estimation performance, but also its good interpretability in revealing the aging discrepancy. In the future, we will consider to extend the proposed framework to race-and-gender-aware age estimation.

\section*{Acknowledgment} \label{sec:acknowledgment}
This work was partially supported by the National Natural Science Foundation of China under Grant $61472186$, the Specialized Research Fund for the Doctoral Program of Higher Education under Grant $20133218110032$, the Funding of Jiangsu Innovation Program for Graduate Education under Grant $CXLX13\_159$, and the Fundamental Research Funds for the Central Universities and Jiangsu \emph{Qing-Lan Project}.



\ifCLASSOPTIONcaptionsoff
  \newpage
\fi



%



\bibliographystyle{IEEEtran}

%

\begin{IEEEbiography}[{\includegraphics[width=1in,height=1.25in,clip,keepaspectratio]{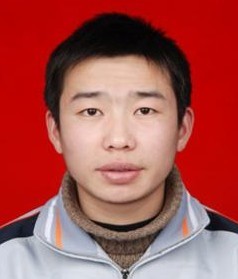}}]{Qing Tian}
received the B.S. degree in computer science from Southwest University for Nationalities, China, and the M.S. degree in computer science from Zhejiang University of Technology, China, respectively with the honor of \emph{Sichuan province-level outstanding graduate} and \emph{Zhejiang province-level outstanding graduate} in 2008 and 2011. From Feb 2011 to Feb 2012, as a researcher in the fields of machine learning and pattern recognition, he worked at ArcSoft, Inc., USA. Since Apr 2012, he has been a Ph.D. candidate in computer science at Nanjing University of Aeronautics and Astronautics, and his current research interests include machine learning and pattern recognition.
\end{IEEEbiography}
\begin{IEEEbiography}[{\includegraphics[width=1in,height=1.25in,clip,keepaspectratio]{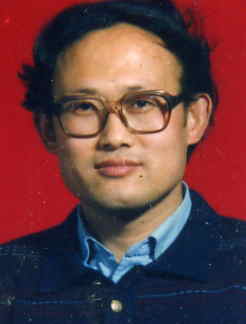}}]{Songcan Chen}
received the B.S. degree from Hangzhou University (now merged into Zhejiang University), the M.S. degree from Shanghai Jiao Tong University and the Ph.D. degree from Nanjing University of Aeronautics and Astronautics (NUAA) in 1983, 1985, and 1997, respectively. He joined in NUAA in 1986, and since 1998, he has been a full-time Professor with the Department of Computer Science and Engineering. He has authored/co-authored over 170 scientific peer-reviewed papers and ever obtained Honorable Mentions of 2006, 2007 and 2010 Best Paper Awards of Pattern Recognition Journal respectively. His current research interests include pattern recognition, machine learning, and neural computing.
\end{IEEEbiography}
\begin{IEEEbiography}[{\includegraphics[width=1in,height=1.25in,clip,keepaspectratio]{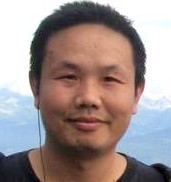}}]{Xiaoyang Tan}
 received his B.S. and M.S. degrees in computer applications from Nanjing University of Aeronautics and Astronautics (NUAA) in 1993 and 1996, respectively. Then he worked at NUAA in June 1996 as an assistant lecturer. He received a Ph.D. degree from Department of Computer Science and Technology of Nanjing University, China, in 2005. From Sept. 2006 to Oct. 2007, he worked as a postdoctoral researcher in the LEAR (Learning and Recognition in Vision) team at INRIA Rhone-Alpes in Grenoble, France. His research interests are in face recognition, machine learning, pattern recognition, and computer vision.
\end{IEEEbiography}







\end{document}